\documentclass[letterpaper, 10 pt, conference]{ieeeconf}  %
\IEEEoverridecommandlockouts                              %
\usepackage{fp}
\newcommand{\pagebudget}[1]{}
\newcommand{\showtotalpagebudget}[1]{}
\usepackage{outlines}
\renewcommand*{\gobble}[1]{}

\usepackage{soul} %
\setstcolor{blue}

\usepackage[dvipsnames]{xcolor}
\usepackage{pgf,tikz,pgfplots}
\PassOptionsToPackage{dvipsnames,table}{xcolor}
\pgfplotsset{compat=1.15}
\usepackage{mathrsfs}
\usetikzlibrary{arrows,backgrounds,shapes}
\usepackage[style=ieee,backend=biber,dashed=false,doi=false,eprint=false, sorting=nyt]{biblatex}
\bibliography{main}
\usepackage{pdfpages}
\usepackage{subcaption}
\usepackage{algorithmicx}
\usepackage{algorithm}
\usepackage[noend]{algpseudocode}
\usepackage{amsmath} 
\usepackage{amssymb} 
\usepackage{array}
\usepackage{epsfig}
\usepackage{enumerate}
\usepackage{graphicx} 
\usepackage{upgreek}
\usepackage[english]{babel}
\usepackage{csquotes}   %
\usepackage[normalem]{ulem}
\usepackage{xspace}
\usepackage{rotating}
\usepackage{tabularx}
\usepackage{pseudocode}
\usepackage{subfloat}
\usepackage{svg}
\usepackage{tikz}

\usepackage[T3,T1]{fontenc}
\DeclareSymbolFont{tipa}{T3}{cmr}{m}{n}
\DeclareMathAccent{\invbreve}{\mathalpha}{tipa}{16}

\algrenewcommand\algorithmicrequire{\textbf{Precondition:}}
\algrenewcommand\algorithmicensure{\textbf{Postcondition:}}
\algnewcommand\algorithmicinput{\textbf{Input:}}
\algnewcommand\Input{\item[\algorithmicinput]}
\algnewcommand\algorithmicoutput{\textbf{Output:}}
\algnewcommand\Output{\item[\algorithmicoutput]}
\newtheorem{theorem}{Theorem}[section]
\newenvironment{definition}[1][Definition]{\begin{trivlist}
\item[\hskip \labelsep {\bfseries #1}]}{\end{trivlist}}
\newcolumntype{C}{>{\centering\arraybackslash} m{2.2in} }

\renewcommand{\Re}{\mathbb{R}}

\newcommand{\old}[1]{{}}
\makeatletter
\newcommand{\Rmnum}[1]{\expandafter\@slowromancap\romannumeral #1@}
\makeatother

\newcommand{\sample}{junction sampling\xspace}

\newcommand{\jpc}{JPC\xspace}
\newcommand{\jpcs}{JPCs\xspace}
\newcommand\mydots{\ifmmode\ldots\else\makebox[1em][c]{.\hfil.\hfil.}\thinspace\fi}

\usepackage{support-caption}   %
\usepackage[font=footnotesize]{caption}%
\makeatletter
\def\thm@space@setup{%
  \thm@preskip=.5pt plus .5pt minus .5pt
  \thm@postskip=\thm@preskip %
}
\makeatother

\title{\LARGE \bf
Rapid Recovery from Robot Failures\\in Multi-Robot Visibility-Based Pursuit-Evasion
}

\author{Trevor Olsen, Nicholas M. Stiffler, and Jason M. O'Kane%
\thanks{%
	The authors are with the Department
    of Computer Science and Engineering, University of South Carolina, Columbia, SC 29208, USA. 
    {\tt \footnotesize \{tvolsen\}@email.sc.edu \{stifflen, jokane\}@cse.sc.edu} %
	This material is based upon work supported by the National Science Foundation 
    under Grant Nos. 1659514 and 1849291.%
}}

\begin{document}

\maketitle
\thispagestyle{empty}
\pagestyle{empty}

\begin{abstract}
This paper addresses the visibility-based pursuit-evasion problem where a team of pursuer robots operating in a two-dimensional polygonal space seek to establish visibility of an arbitrarily fast evader. 
This is a computationally challenging task for which the best known complete algorithm takes time doubly exponential in the number of robots. However, recent advances that utilize sampling-based methods have shown progress in generating feasible solutions.
An aspect of this problem that has yet to be explored concerns how to ensure that the robots can recover from catastrophic failures which leave one or more robots unexpectedly incapable of continuing to contribute to the pursuit of the evader. 
To address this issue, we propose an algorithm 
that can rapidly recover from catastrophic failures. When such failures occur, a replanning occurs, leveraging both the information retained from the previous iteration and the partial progress of the search completed before the failure to generate a new motion strategy for the reduced team of pursuers.
We describe an implementation of this algorithm and provide quantitative results that show that the proposed method is able to recover from robot failures more rapidly than a baseline approach that plans from scratch.
\end{abstract}

\section{Introduction}
For a number of important applications of mobile robots, including
  environmental monitoring~\cite{IslNoo+15,DamTokKum17,TokBha+10},
  disaster recovery~\cite{HolKehSin07}, and
  surveillance~\cite{AceBeg+14},
the central problem is to plan motions for a team robots called \emph{pursuers} to locate unpredictably moving agents called \emph{evaders}.
Though progress has been made on several forms of this problem, a key limitation within much of the existing work is an inability to adapt in cases of unrecoverable failures of individual robots. 
Such failures are particularly likely to occur in the very application domains for which these methods are well-suited.

\begin{figure}
    \includegraphics[width=0.48\columnwidth]{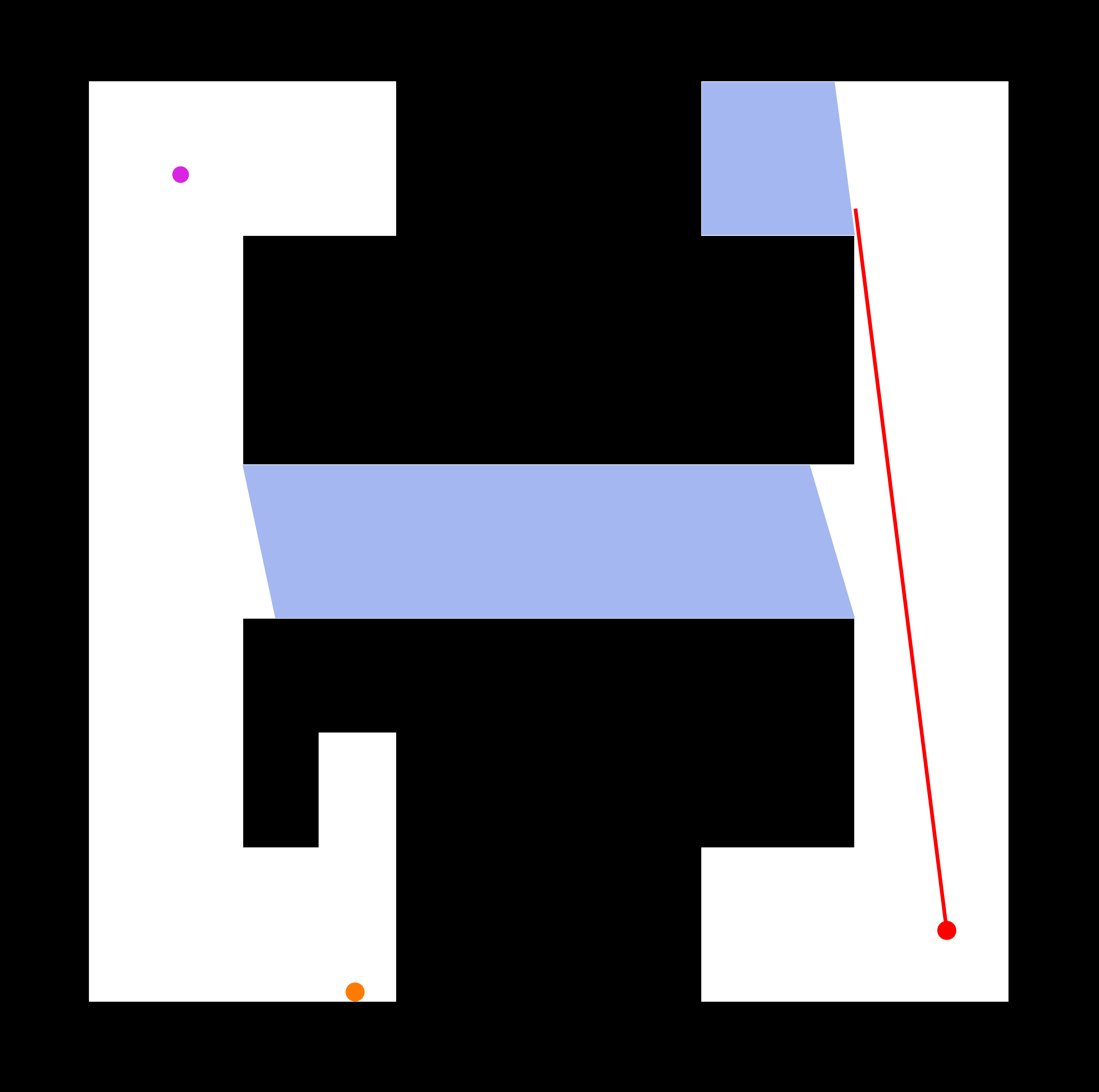}\hfill
    \includegraphics[width=0.48\columnwidth]{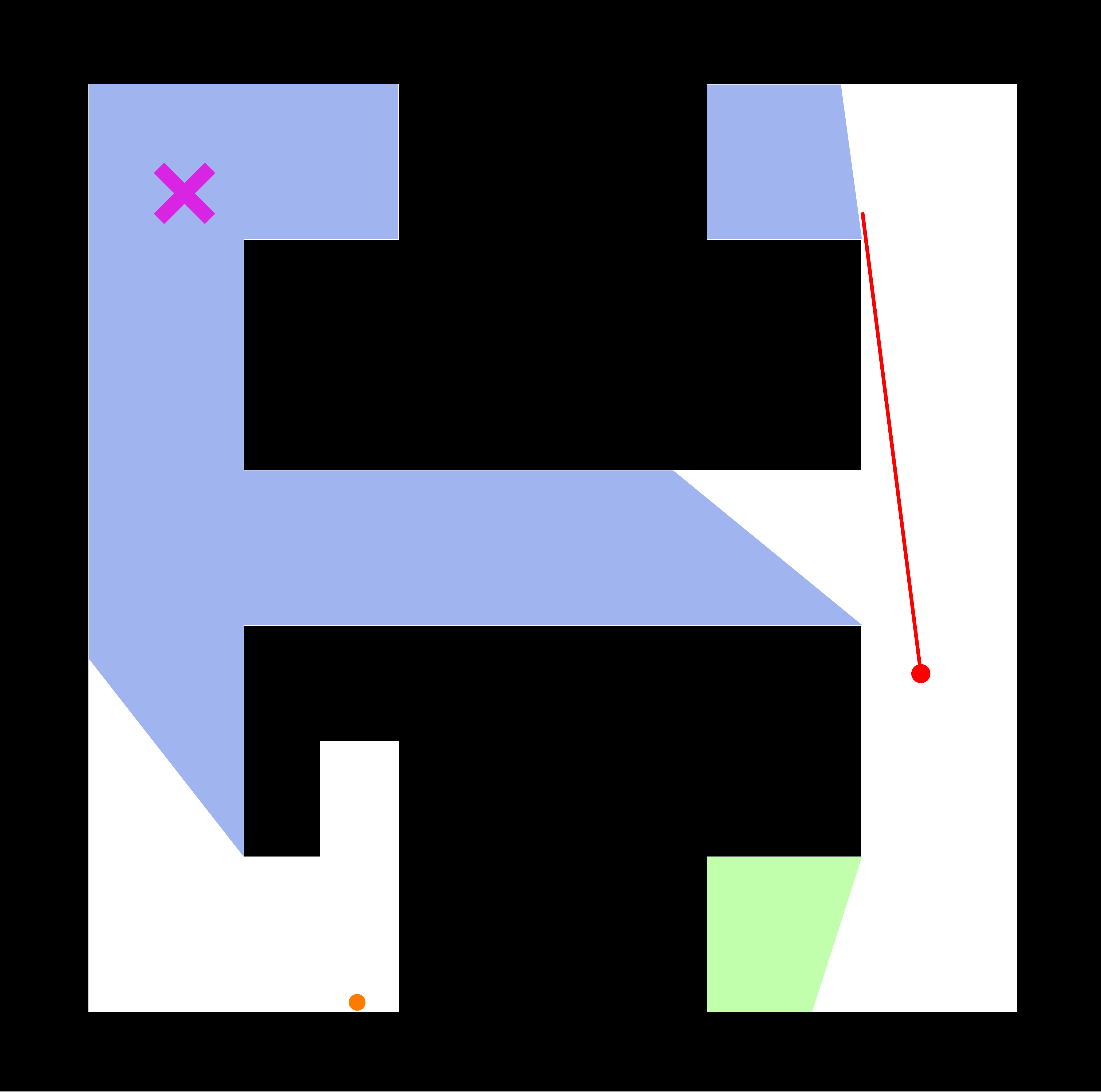}
    \caption{[left] An initial joint plan for 3 robots to search a simple environment for an evader. Two robots on the left remain stationary and monitor that side of the environment while a third moves to search the right.  [right] The robot in the top left corner fails unexpectedly.  As a result, the initial plan is no longer correct, because it cannot locate an evader that moves into the top left corner of the environment.}\label{fig:page1fig}
\end{figure}

Figure~\ref{fig:page1fig} provides a simple illustration of this problem, in a context wherein each pursuer is equipped with an omnidirectional sensor that can detect the evaders within its line of sight.  The objective is to move the pursuers along paths that are guaranteed to locate the evaders, even if the evaders may move arbitrarily quickly. An initial plan to search this environment using three robots appears on the left of Figure~\ref{fig:page1fig}.
Now suppose the robot stationed at the top left of the environment fails during the plan's execution, as shown on the right of Figure~\ref{fig:page1fig}.  In this case, any evader hiding in the center of the environment at that time can now escape to the top left corner without being detected.  Thus, the initial plan is no longer guaranteed to locate the evader. 

\begin{figure}
    \includegraphics[width=0.48\columnwidth]{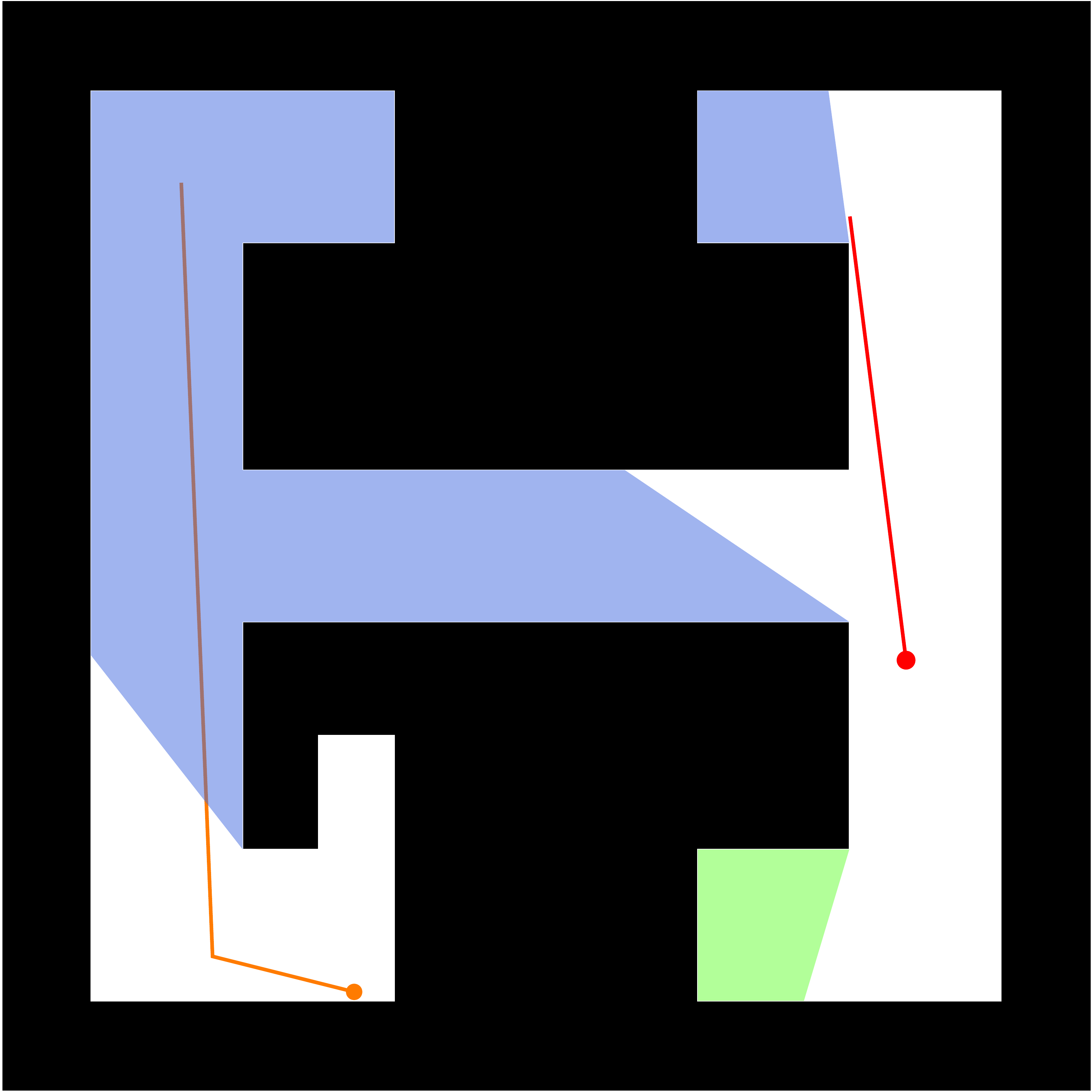}\hfill
    \includegraphics[width=0.48\columnwidth]{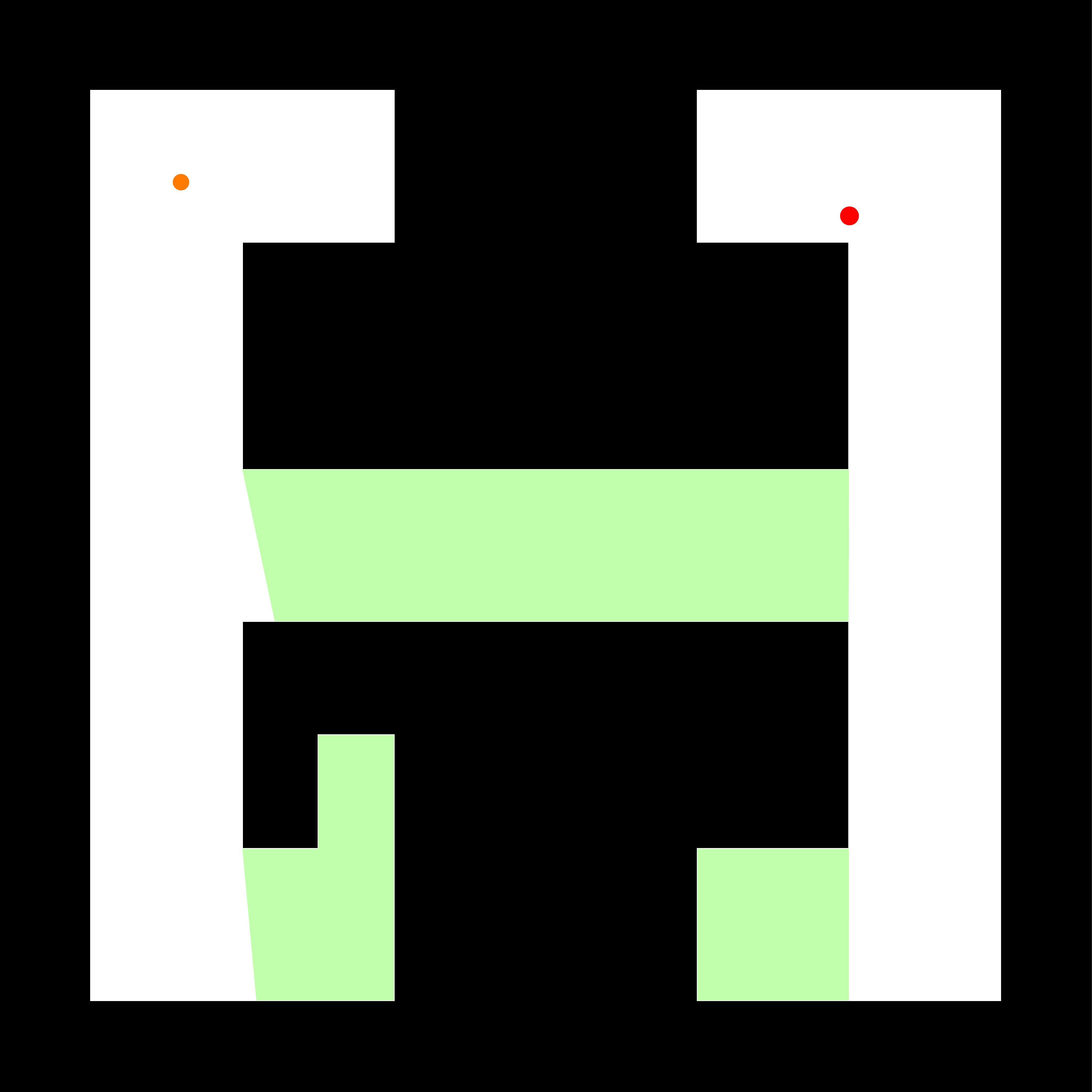}
    \caption{[left] The robots from Figure~\ref{fig:page1fig} replan, using the proposed algorithm, to form a correct plan for the remaining 2 robots. [right] The plan completes successfully.}
    \label{fig:page1fig2}
\end{figure}

In response to this limitation, this paper proposes a new approach for this form of visibility-based pursuit-evasion problem, suitable for contexts in which failures of robots are likely.  Because generating pursuit plans for this domain is computationally challenging~\cite{StiOKa14a}, we introduce a method for replanning in cases of robot failure that leverages information derived from the previous plan to accelerate the construction of new plans for the unexpectedly-smaller teams. Figure~\ref{fig:page1fig2} shows an example of recovery from the failure depicted in Figure~\ref{fig:page1fig}.

The central idea is to adapt known methods for solving these problems in the failure-free case, which are based on a roadmap-like data structure called the sample-generated pursuit-evasion graph (SG-PEG)~\cite{StiOKa14b}.  For planning the motions of $n$ robots in this pursuit-evasion problem, the SG-PEG graph represents the connectivity of the joint configurations space with vertices that represent joint configurations and edges that represent collision free movements, much like a traditional probabilistic roadmap.  In addition, each vertex is labeled with information that encodes which hidden portions of the environment can be cleared of unseen evaders along some walk in the graph ending at that vertex.  The new replanning approach modifies this data structure when a robot fails to reflect the removal of that robot from the search.  The remaining graph then provides a valuable starting point for the process of planning to solve the problem with the remaining $n-1$ robots.  The algorithm then expands this graph by adding additional vertices, attempting to recover the contributions made by the failed robot.

This work is, to the authors' best knowledge, the first to address the problem of recovery from pursuer failures for this form of pursuit-evasion problem. 
In the remainder of the paper, we review some related work (Section~\ref{sec:related}), formally define the problem (Section~\ref{sec:ps}).  Then we describe and evaluate the algorithm (Sections~\ref{sec:ad} and \ref{sec:experiments} respectively) before concluding with discussion and a preview of future work (Section~\ref{sec:conclusion}).

\section{Related work}\label{sec:related}
The visibility-based pursuit-evasion problem posed in this paper can be thought of as a specialization
of the broader problem of \textsl{search and target tracking}. The common theme across this work is
the pursuit of an agent (or agents) by one (or more) pursuers to either establish or maintain visibility of the target~\cite{SkhDud17,SkhKakDud18,ZouBha19,}.

The literature on these problems %
can by understood by organizing
according to underlying models, including differential game theory, graph variants, and geometric variants. 
Though this work considers the geometric formulation, we present a brief synopsis of the contributions in the domains of differential game theory and graph theory.

The seminal work of Isaacs~\cite{Isa65} and Ho et. al.~\cite{HoBryBar65} was the first to adapt the pursuit-evasion problem to a dynamic game-theoretic framework. This remains an active research area~\cite{CheZha+16,KumLuMPat17,RuiMur13,VanIsl14}.
Recent results include continued progress by  
utilizing techniques such as reinforcement learning~\cite{WanDonSun20} and the exploitation of rich representations such as Voronoi partitions to aid in the search~\cite{MolCas+18,ZhoZha+16}.

A different formulation in which the domain is modeled as a discrete graph
was initially proposed by Parsons~\cite{Par76} and is referred to as the edge-searching problem. Petrov
later 
independently rediscovered some of Parsons' results in the context of
differential
game theory~\cite{Pet83}. Golovach later showed that both problems considered an equivalent discrete game on graphs~\cite{Gol89}.
A number of survey papers~\cite{AbrPet2013,Als04,BorKoeTov13} provide overviews of the many problem variants 
that can be realized within the graph model, such as specifying the rules of movement for the pursuers and for the evaders~\cite{SkhDud13}, the kind of graph~\cite{KolCar10}, etc.

This paper specifically focuses on a variant of the problem where the pursuers and evaders operate in a geometric environment~\cite{GuiLat+99,ParLeeChwa01, SuzYam92}.
There are a number of results for the single pursuer variant of the problem that range 
from providing theoretical properties such as completeness~\cite{GuiLat+99} and optimality~\cite{StiOKa17},
to more restricted scenarios where there are limits on the actuation and sensing capabilities of the 
pursuers \cite{,BhaKle+12,LiAmi+18,StiOKa20,TovLav08}.
Due to the broad range of practical applications, the multi-pursuer variant of the problem
has drawn continued interest \cite{DurFraBul12, GreGiv+17, KolCar10a}, \cite{StiOKa14b} in recent years.
The multi-pursuer scenario poses additional challenges owing to the problem complexity~\cite{StiOKa14a}.
A common thread through much of the existing work is an assumption that the
pursuer(s) can reliably execute the trajectories generated for them by the
planner. This work seeks to address this limitation.

\section{Problem statement}\label{sec:ps}

We first define the basic problem in the absence of pursuer failures (Section~\ref{ss:notation}) and describe how to cast that problem in terms of a discrete
representation of which areas of the environment are `clear' or `contaminated' (Section~\ref{ss:background}), before introducing the possibility of pursuer failures (Section~\ref{ss:fail}).

\subsection{Environment, pursuers, and evaders}\label{ss:notation} \pagebudget{0.6}

 The \emph{environment} $F$ is a closed, bounded, and connected polygonal region in $\Re^2$.
A team of $n$ \emph{pursuers}, who can travel throughout the environment at bounded speed, are equipped with omnidirectional sensors whose range is only bounded by line of sight within the environment. That is, a pursuer at point $q \in F$ can detect anything within its \emph{visibility polygon} $V(q) = \{r \in F \mid \overline{qr} \subset F \}$. We  denote the location of the $i^{\text{th}}$ pursuer as a function of the time $t$ by the continuous function $f_i(t):[0, T] \rightarrow F $, in which $T$ is some termination time which the pursuers may choose. The $n$-robot \emph{joint pursuer configuration} (JPC) at time $t$ is the vector $\langle f_1(t), f_2(t), \mydots, f_n(t) \rangle \in F^{n}$.

A single \emph{evader} seeks to avoid detection by the pursuers by moving continuously within the environment, without any bound on its speed. We denote its location, as a function of time, by the continuous function $e(t): [0, \infty) \rightarrow F$, unknown to the pursuers.
Observe that, because we plan for the worst case, any strategy for the pursuers that guarantees detection of a single evader can also guarantee detection for each of potentially many evaders.

\begin{figure}
    \centering
    \resizebox{0.95\columnwidth}{!}{
    \begin{tikzpicture}[scale=0.75, yscale=0.75]
        \fill[draw=black,fill=black] (-1,-1) rectangle (15,12);
        \fill[color=white, line width=0.3mm] (0,0)--(0,3)--(1.5,3)--(1.5,4)--(0,4)--(0,7)--(1.5,7)--(1.5,8)--(0,8)--(0,11)--(4,11)--(4,8)--(2.5,8)--(2.5,7)--(4,7)--(4,6)--(5,6)--(5,7)--(6.5,7)--(6.5,8)--(5,8)--(5,11)--(9,11)--(9,10)--(10,10)--(10,11)--(14,11)--(14,8)--(10,8)--(10,9)--(9,9)--(9,8)--(7.5,8)--(7.5,7)--(9,7)--(9,6)--(10,6)--(10,7)--(14,7)--(14,4)--(12.5,4)--(12.5,3)--(14,3)--(14,0)--(10,0)--(10,1)--(9,1)--(9,0)--(5,0)--(5,1)--(4,1)--(4,0);
        \fill[color=black] (11.5,4)--(10,4)--(10,5)--(9,5)--(9,4)--(5,4)--(5,5)--(4,5)--(4,4)--(2.5,4)--(2.5,3)--(4,3)--(4,2)--(5,2)--(5,3)--(9,3)--(9,2)--(10,2)--(10,3)--(11.5,3);

        \draw [cyan] plot coordinates {(2,3.5) (2,9.5)};
        \draw [cyan] plot coordinates {(4.5, 1.5) (7, 1.5)};
        \draw [cyan] plot [smooth, tension=1] coordinates {(2,3.5) (2.75,1.75) (4.5,1.5)};
        
        \draw [WildStrawberry] plot coordinates {(12, 1.5) (12, 3.5)};
        \draw [WildStrawberry] plot [smooth, tension=1] coordinates {(12,3.5) (11.25,5) (9.5,5.5)};
        \draw [WildStrawberry] plot [smooth, tension=1] coordinates {(7,7.5) (7.75,5.75) (9.5,5.5)};
        \draw [WildStrawberry] plot [smooth, tension=1] coordinates {(7,7.5) (7.75,9.25) (9.5,9.5)};
        \draw [WildStrawberry] plot coordinates {(9.5,9.5) (11,9.5)};
     
        \node[fill=cyan, circle, draw=black, scale=0.5, label=right:{$f_1(0)$}] (p1) at (7,1.5) {};
        \node[fill=cyan, circle, draw=black, scale=0.5, label=left:{$f_1(t)$}] (pt) at (2.35, 2.25) {};
        \node[fill=cyan, circle, draw=black, scale=0.5, label=above:{$f_1(T)$}] (p2) at (2,9.5) {};
        \node[fill=WildStrawberry, circle, draw=black, scale=0.5, label=below:{$f_2(0)$}] (q1) at (12,1.5) {};
        \node[fill=WildStrawberry, circle, draw=black, scale=0.5, label=right:{$f_2(t)$}] (qt) at (7.5,6) {};
        \node[fill=WildStrawberry, circle, draw=black, scale=0.5, label=above:{$f_2(T)$}] (q2) at (11,9.5) {};
        \node[fill=Fuchsia, circle, draw=black, scale=0.5, label=above:{$e(0)$}] (e1) at (3,4.5) {};        
        \node[fill=Fuchsia, circle, draw=black, scale=0.5, label=left:{$e(t)$}] (et) at (6.5,6.5) {};
        \node[fill=Fuchsia, circle, draw=black, scale=0.5, label=right:{$e(T)$}] (ett) at (5.5,10) {};
        \draw [Fuchsia] plot [smooth, tension=1] coordinates {(e1) (4.5,5.5) (7,5.75) (6,5)  (6.75, 8) (ett)};
        \draw [color=black,dashed] (qt)--(et);
    \end{tikzpicture}}
    \caption{Two pursuers
    execute a search. Note that at time $t$, the pursuer on path $f_{2}$ is capable of detecting the evader $e$.} 
    \label{fig:notation}
\end{figure}
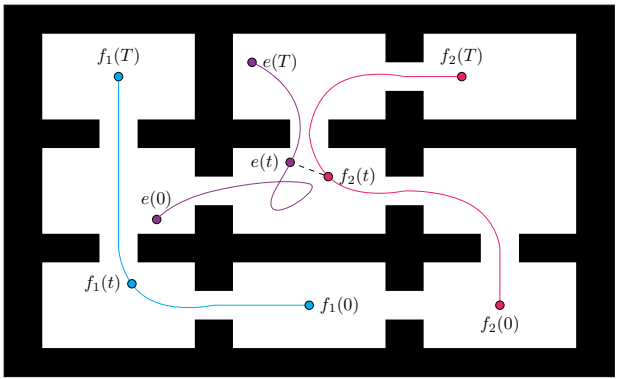

The pursuers' objective is to establish visibility with the evader.  Thus, for a given environment $F$, the goal is to choose the termination time $T$ and the functions $f_1, f_2, \mydots, f_n$ to ensure that
  for any evader trajectory $e$, 
  there exists a time $t_0 \in [0, T]$,
  such that
  $e(t_0) \in \bigcup_{i \leq n} V(f_i(t_0)) $.
  Figure~\ref{fig:notation} illustrates the notation.

\subsection{Shadows}\label{ss:background}
The primary difficulty in this type of visibility-based pursuit-evasion concerns reasoning about the regions 
of the environment that are not currently visible to the pursuers at the present time.
To resolve that difficulty, Guibas, Latombe, LaValle, Lin, and Motwani \cite{GuiLat+99} introduced a reformulation of the
problem, based upon tracking which, if any, of the regions of the environment not currently perceptible by the pursuers might contain an as-yet-undetected evader.

To formalize this idea, define the \emph{shadow region}, $S(t) = F \setminus \bigcup_{i \leq n} V(f_i(t)) $ as the portion of the environment unseen by any pursuer at time $t$. The maximal connected components of $S(t)$ are called \emph{shadows}.
The terms \emph{cleared} and \emph{contaminated} can be applied to a shadow to reflect the relative
\emph{status} of the shadow at that point in the pursuers' search. A cleared shadow is an unseen area of the environment that, based upon the pursuers' motions up to the current time $t$, is guaranteed to not contain an unseen evader. Any shadow that is not cleared is called contaminated.

To compactly describe the status for all of the shadows, we utilize a \emph{shadow label}, which is a binary string comprised of one bit for each shadow, in which the $i^{\text{th}}$ bit is $1$ if the $i^{\text{th}}$ shadow (in an arbitrary but fixed ordering) is contaminated, and $0$ otherwise.

Though shadows change continuously as the pursuers move within $F$, the cardinality of the shadows and their labels can change only when a shadow event occurs, i.e. a shadow appears, disappears, splits, or multiple shadows merge into a single shadow.

\begin{itemize}
    \item \emph{Appear}: A shadow appears when the pursuers move in such a way that makes a previously observed part of the environment no longer visible. In this case, the new shadow is assigned a cleared status.
    \item \emph{Disappear}: If a pursuer gains vision of a shadow, we say the shadow disappears. Here, the shadow bit is deleted.
    \item \emph{Split}: If a pursuers' vision disconnects a shadow, we say that the shadow has split. Each new component is given the label of the original pre-split shadow.
    \item \emph{Merge}: If two or more shadows become a single connected component, we say the shadows have merged. The newly merged shadow takes on the cleared status if and only if each merging component was cleared prior to the merge. Otherwise, the merged shadow is considered to be contaminated.
\end{itemize}

\begin{figure}
  \begin{subfigure}[t]{.49\columnwidth}
    \centering
    \includegraphics[width=1\columnwidth]{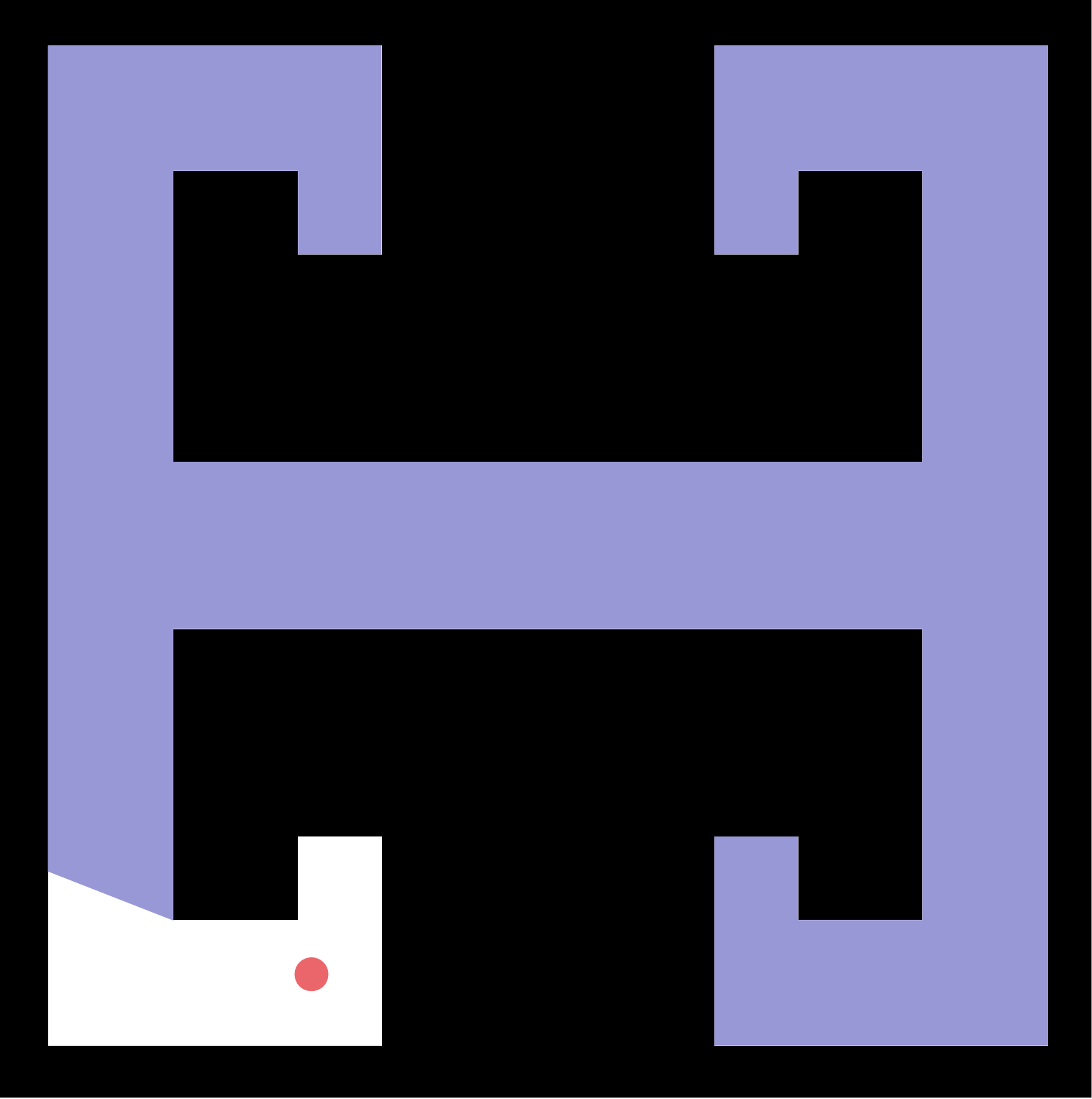}
    \caption{The initial location of a single pursuer. Here, the shadow label is 1.}
  \end{subfigure}
  \begin{subfigure}[t]{.49\columnwidth}
    \centering
    \includegraphics[width=1\columnwidth, height=4.26cm]{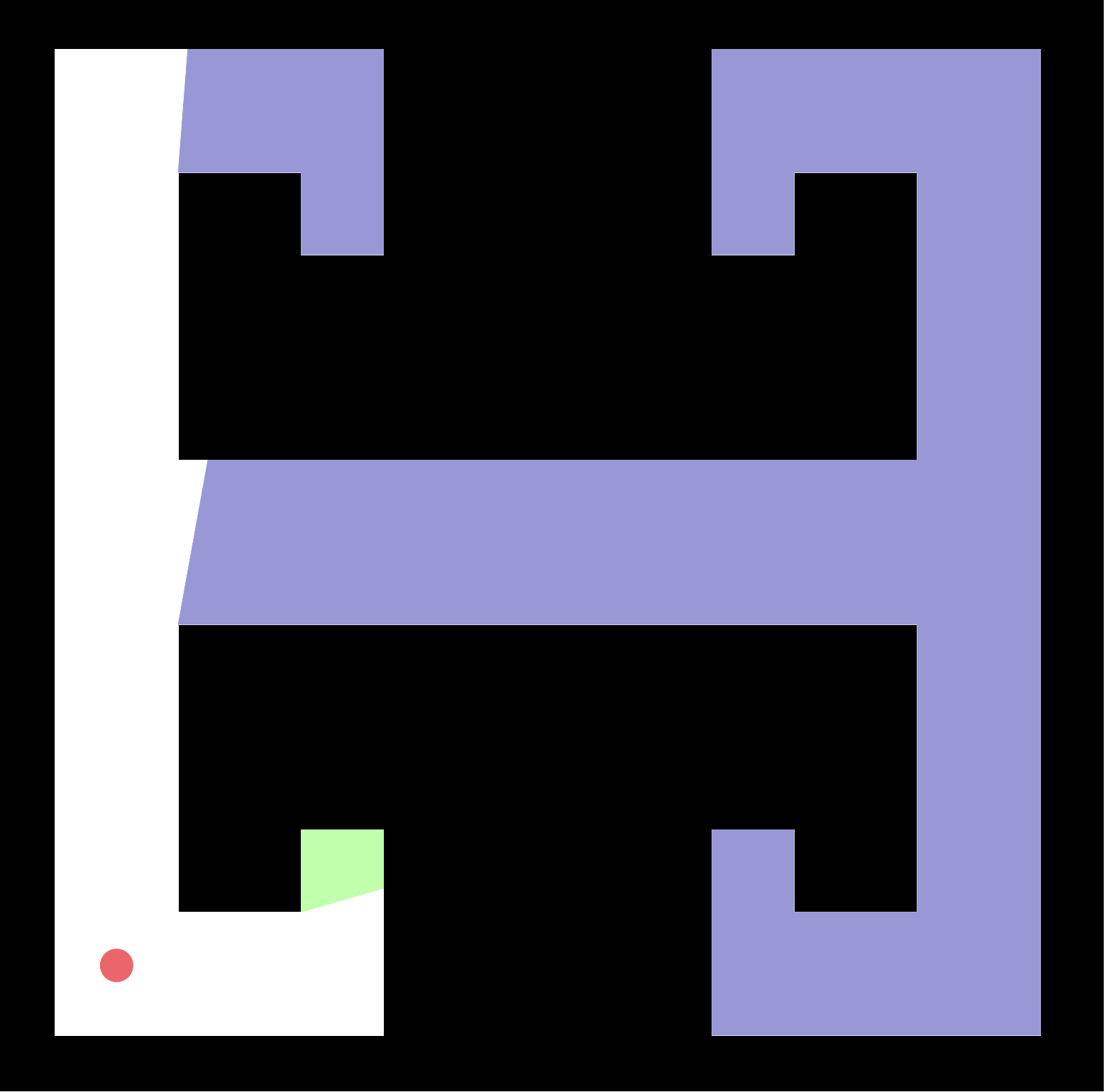}
    \caption{Movement of a pursuer that would cause one shadow to appear and another to split. The result has one cleared and two contaminated shadows.  In our implementation, this scenario is assigned shadow label 110.}
  \end{subfigure}
  \caption{An example of shadow events and labels.}
\label{fig:shadows}
\end{figure}

This formulation of the problem in terms of clear and contaminated shadows is valuable because it enables a planner to reason over those shadows and labels, rather than directly over the space of all possible evader paths.
That is, the overall visibility-based pursuit-evasion problem can be restated as a search for pursuer motions that lead to a system state in which the binary string of shadow labels contains all zeroes, indicating that every shadow is clear.

\subsection{Pursuer Failures}\label{ss:fail}

The basic problem described so far has been addressed in prior work in a number of different ways.  The new contribution in this paper is to consider this problem in an online setting, in which the pursuer robots may fail during the execution of a plan.  Such failures are assumed to be both unpredictable and permanent, but known to all of the pursuers when they occur.

\newcommand{\sadt}{{\invbreve{t}}}

More precisely, we consider the case in which the $n$ pursuers are executing the paths
$f_i: [0, T] \to F$ for $i \in \{1,\ldots,n\}$, with termination time $T$.  Suppose that at some time $0 \le \sadt < T$, the $k^{\rm th}$ pursuer fails.
The \emph{replanning problem} addressed in this paper is to generate new pursuer paths
$f_i': [\sadt, T'] \to F$ for $i \in \{1,\ldots,n\} \setminus \{ k \}$.
These revised paths must begin at the pursuers' locations at the time of failure, so that for each surviving pursuer, we have $f_i'(\sadt) = f_i(\sadt)$.  The revised paths end at a new termination time $T'$.

Generalizing the objective from the original failureless model, we say that the pursuers' execution of the prefix of $f_1,\ldots,f_n$ up to time $\sadt$, followed by $f_1',\ldots,f_n'$ from time $\sadt$ to $T'$ is a \emph{solution} if,  for any evader trajectory $e$, either
        (i) there exists a time $t_0 \in [0, \sadt]$, such that $e(t_0) \in \bigcup_{i \leq n} V(f_i(t_0))$, or
        (ii) there exists a time $t_0 \in [\sadt, T']$, such that $e(t_0) \in \bigcup_{i \leq n; i \neq k} V(f_i'(t_0))$.
That is, we seek to guarantee that the evader is seen by any of the robots at some time before or after the pursuer failure, or by one of the non-failing robots at some time after the failure.
Similar but notationally tedious generalizations can be made for multiple failures within a single execution.

Fortunately, we can also generalize the notions of shadow label updates to account for the abrupt change in shadows that occur at time $\sadt$.  Specifically, a shadow extant immediately after a robot failure is contaminated if and only if it intersects with a contaminated shadow from immediately before the failure occurred.  Notice, for example, in the right portion of Figure~\ref{fig:page1fig}, that the large shadow encompassing the center and upper left portion of the environment is marked contaminated because it overlaps the central shadow which was contaminated before the failure.  In contrast, the smaller shadow in the lower right has a clear label after the failure, because the only pre-failure shadow with which it intersects (namely, itself) had a clear label.
This feature of the definition of success, which allows shadows to remain clear even across a failure of one of the pursuers, is crucial because it allows the pursuers the possibility of retaining some of their progress (i.e. cleared shadows) toward completing the task, rather than starting from scratch each time.

\section{Algorithm Overview}\label{sec:ad}
This section provides a detailed description of our algorithm.  Because no efficient algorithm for solving even the failure-free case is known~\cite{StiOKa14a}, we take a sampling-based approach.  The basic idea is to construct a roadmap within the pursuers' joint configuration space, using an existing data structure called the sample-generated pursuit-evasion graph (SG-PEG), which a subset of the present authors originally introduced for the failure-free case~\cite{StiOKa14b}.  We leverage this data structure in a new way by introducing new sampling strategies designed to rapidly re-acquire a solution in cases where a pursuer must be removed.

The core of the algorithm is a method called \textsc{DropRobot} which, given a solution path for $k$ robots (for some $k$), uses an SG-PEG to attempt to rapidly generate a solution for $k-1$ robots, using the original $k$-robot solution as a guide.  Our algorithm relies upon \textsc{DropRobot} both to generate an initial solution for the full set of $n$ robots ---by iteratively reducing from a rapidly-generated trivial solution--- and for replanning when a pursuer fails.

The remainder of this section presents details of the method.  After a brief review of the SG-PEG (Section~\ref{sec:sgpeg}), we describe the \textsc{DropRobot} method (Section~\ref{sec:droprobot}) and how that method is used to generate the initial solution (Section~\ref{sec:initial}) and for replanning (Section~\ref{sec:replan}).

\subsection{SG-PEG}\label{sec:sgpeg}

The SG-PEG is a data structure the represents a roadmap of valid joint paths for a team of pursuers in a known environment $F$, augmented with information about the shadow labels that can be achieved by executing those paths.  We present here a concise overview; additional detail may be found in the original paper~\cite{StiOKa14b}.

An SG-PEG is a directed graph $G=(V_{G},E_{G})$, in which one vertex $v_0$ is designated as the \emph{root vertex}.  Each SG-PEG is constructed for a specific number $n$ of pursuers.
Each vertex $v \in V_{G}$ corresponds to a specific \jpc $\langle p_{1},\ldots,p_{n} \rangle \in F^{n}$.
Each directed edge $e \in E_{G}$ connects two vertices $v, u \in V_{G}$ for which it is possible for every pursuer to make a collision-free straight line motion between the representative configurations. That is, the existence of an edge from $v$ to $u$ means that, for each $1 \le i \le n$, $\overline{v_{i}u_{i}}\subset F$.

In addition to this graph structure, each vertex $v$ maintains a set of reachable shadow labels.  Specifically, a shadow label $\ell$ will be recorded at a particular vertex $v$ as a reachable shadow label if there exists a walk from $v_0$ to $v$ that results in the shadow marked clear within $\ell$ indeed being clear.

The primary operation that can be performed on a SG-PEG is $\textsc{AddSample}(\langle p_{1}, \ldots, p_{n} \rangle)$, which accepts a collision-free \jpc as input and performs the following steps:
\begin{enumerate}[(i)]
    \item It inserts a new vertex $v$ at the given \jpc.
    \item For every existing vertex $u$ for which the segment $\overline{uv}$ is collision free in $F^n$, it adds the edges $\overrightarrow{uv}$ and $\overrightarrow{vu}$.  The operation then computes a mapping that describes how the shadows at vertex $u$ evolve as the pursuers move from the \jpc at vertex $u$ to the \jpc at vertex $v$.  (The inverse mapping is applied to $\overrightarrow{vu}$). \item Finally, the reachable shadow label information across the graph is updated by propagating the reachable shadow labels, using the mappings attached to each edge, recursively across the graph, to determine what new reachable shadow labels, if any, arise due to the inclusion of the new sample $v$.  
\end{enumerate}
The SG-PEG data structure is useful for our problem because, starting from a root vertex at the pursuers' initial positions, executing a sequence of \textsc{AddSample} operations can eventually lead to a vertex being marked with an all-zero reachable shadow label.  From there, a sequence of \jpc{}s solving the problem can readily be extracted by walking backward along through the graph.

\subsection{Dropping a robot}\label{sec:droprobot}

Suppose $k$ pursuers are at some \jpc $q$ with shadow label $\ell$, and have computed a sequence of future \jpcs to visit that will solve the problem from that point, eventually reaching \jpc with an all-clear shadow label.  How can we use this information to construct a new solution that can be executed from this point by only $k-1$ of these pursuers, removing one particular pursuer from the solution?
Notice that this scenario applies both to the case of a failed pursuer (in which case $q$ and $\ell$ can be derived from the current state when the failure occurred, and $\ell$ may mark some shadows as clear) and to a complete solution starting from the pursuers' starting position and all-contaminated shadow label.
To simplify the notation below, we assume without loss of generality that $n^{\text{th}}$ pursuer is the one removed.

The \textsc{DropRobot} method, shown in Algorithm~\ref{algo:drop}, solves this problem.
The algorithm constructs an SG-PEG $G_{k-1}$, starting with a root vertex at which the $n^{\text{th}}$ pursuer has been removed and the shadow label has been updated accordingly.  From there, it adds a collection of \emph{junction samples}, designed to recover information lost due to the removal of the $n^{\text{th}}$ pursuer at each step of the existing solution.  If $G_{k-1}$ does not contain a solution after that step, \textsc{DropRobot} continues by inserting additional samples called \emph{web samples} designed to provide good coverage, in the sense of visibility, of the environment.  The process continues until a solution is found, or until some arbitrary timeout expires.  Details about junction sampling and web sampling appear below.

\begin{algorithm}[t]
    \caption{\textsc{DropRobot}($F, k, q_1,\ldots, q_m, \ell$)}
    \label{algo:drop}
    \begin{algorithmic}[1]
        \Input{An environment $F$; a positive integer $k$; a sequence $q_1, \ldots, q_m$ of $k$-pursuer \jpcs ; a shadow label $\ell$ for $q_1$.}
        \Output{A sequence $q'_1, \ldots, q'_{m'}$ of $(k-1)$-pursuer \jpcs leading to an all clear shadow label at $q'_{m'}$ or \textsc{Failed}.}
        
        \State{$G_{k-1} \gets $ new SG-PEG for $k-1$ pursuers}
        \State{$\langle p_1, \ldots, p_k \rangle \gets q_1$}
        \State{$r \gets G_{k-1}.\textsc{AddRoot}(\langle p_1, \ldots, p_{k-1} \rangle)$}
        \State{$\ell' \gets \ell$ updated for the removal of $p_n$}
        \State{$r.\textsc{AddReachable}(\ell')$}
        \For{$i \gets 1, \ldots, m$}
            \State{\textsc{AddJunctionSamples}($k$, $G_{k-1}$, $q_i$)}
        \EndFor
        
        \While{$G_{k-1}$ has no solution \textbf{and} time remains}
            \State{$q \gets $ \textsc{WebSample}($G_{k-1}$)}\label{line:web}
            \State{$G.\textsc{addSample}(q)$}
        \EndWhile
        \If{$G_{k-1}$ has a solution}
            \State{\Return{$G_{k-1}$.\textsc{ExtractSolution}()}}
        \Else
            \State{\Return{\textsc{Failed}}}
        \EndIf
    \end{algorithmic}
\end{algorithm}

\begin{algorithm}[t]
  \caption{\textsc{AddJunctionSamples}($k$, $G_{k-1}$, $q$)} 
  \label{algo:junction}
  \begin{algorithmic}[1]
    \Input{A positive integer $k$; an SG-PEG $G_{k-1}$ for $k-1$ pursuers;
    a $k$-pursuer \jpc $q$}
    \Output{No return value, but samples are added to $G_{k-1}$.}

    \State{$\langle p_1, \ldots, p_k \rangle \gets q$}
    \State{$G_{k-1}.\textsc{addSample}(\langle p_1, \mydots, p_{n-1} \rangle)$}\label{line:jpc1}
    \For{$i \gets 1,\ldots,n-1$}
        \If{$V(\textcolor{purple}{p_i}) \cap V(p_n) \ne \emptyset$}
            \State $z \gets $ random point in $V(\textcolor{purple}{p_i}) \cap V(p_n)$
            \State $G.\textsc{addSample}(\langle p_1, p_2, \mydots, \textcolor{purple}{z}, \mydots, p_{n-1} \rangle)$\label{line:jpc3}
            \State $G.\textsc{addSample}(\langle p_1, p_2, \mydots, \textcolor{purple}{p_n}, \mydots, p_{n-1} \rangle)$\label{line:jpc2}

        \EndIf
    \EndFor
  \end{algorithmic}
\end{algorithm}

\subsubsection{Junction sampling}\label{ssect:js}

\begin{figure}
  \begin{subfigure}[t]{.49\columnwidth}
    \centering
    \includegraphics[width=1\columnwidth]{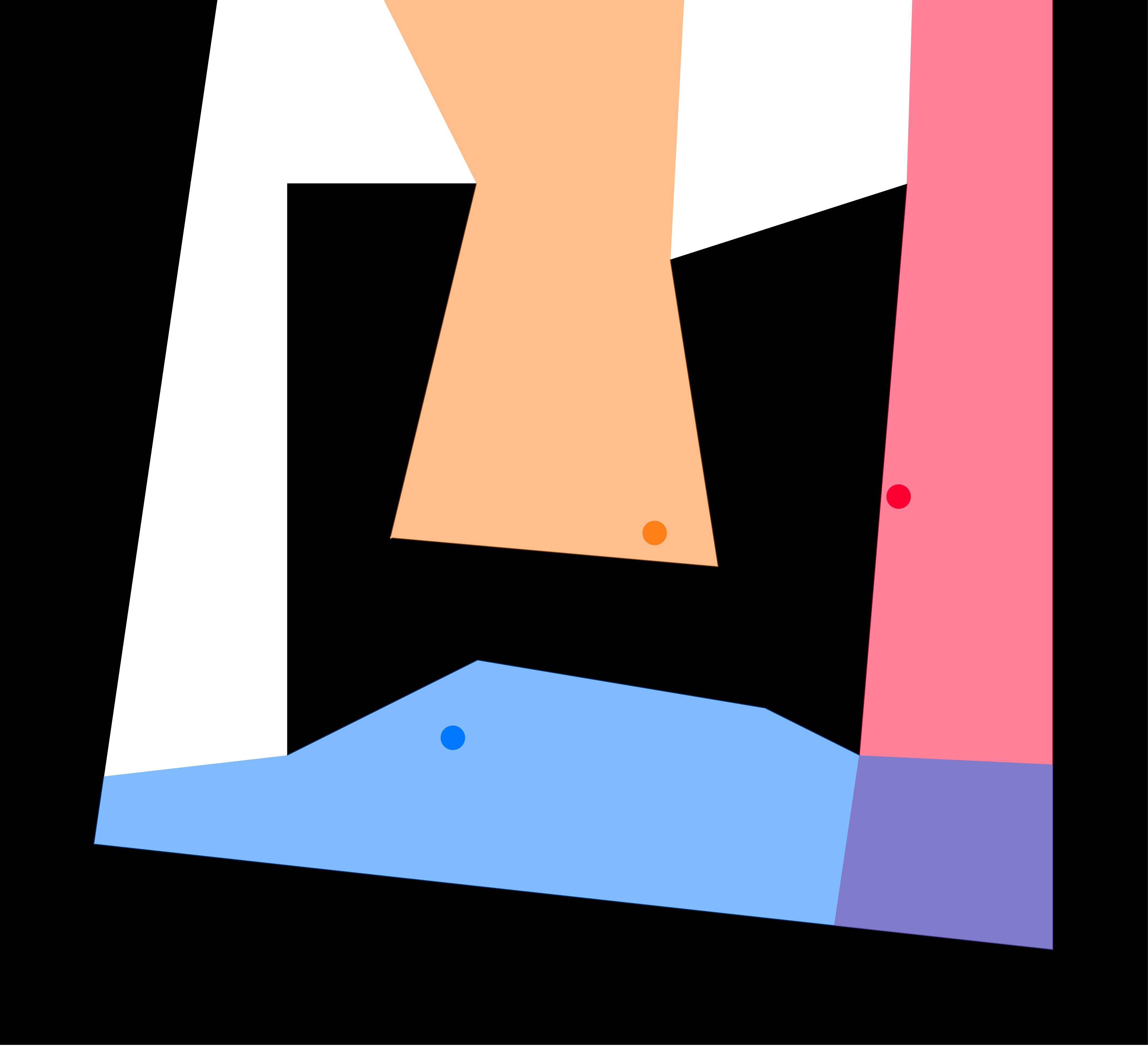}
    \caption{The initial JCP. The \textcolor{blue}{$n^{\text{th}}$ pursuer is blue} [bottom], and the \textcolor{purple}{$i^{\text{th}}$ pursuer is red} [right].}
  \end{subfigure}
  \begin{subfigure}[t]{.49\columnwidth}
    \centering
    \includegraphics[width=1\columnwidth]{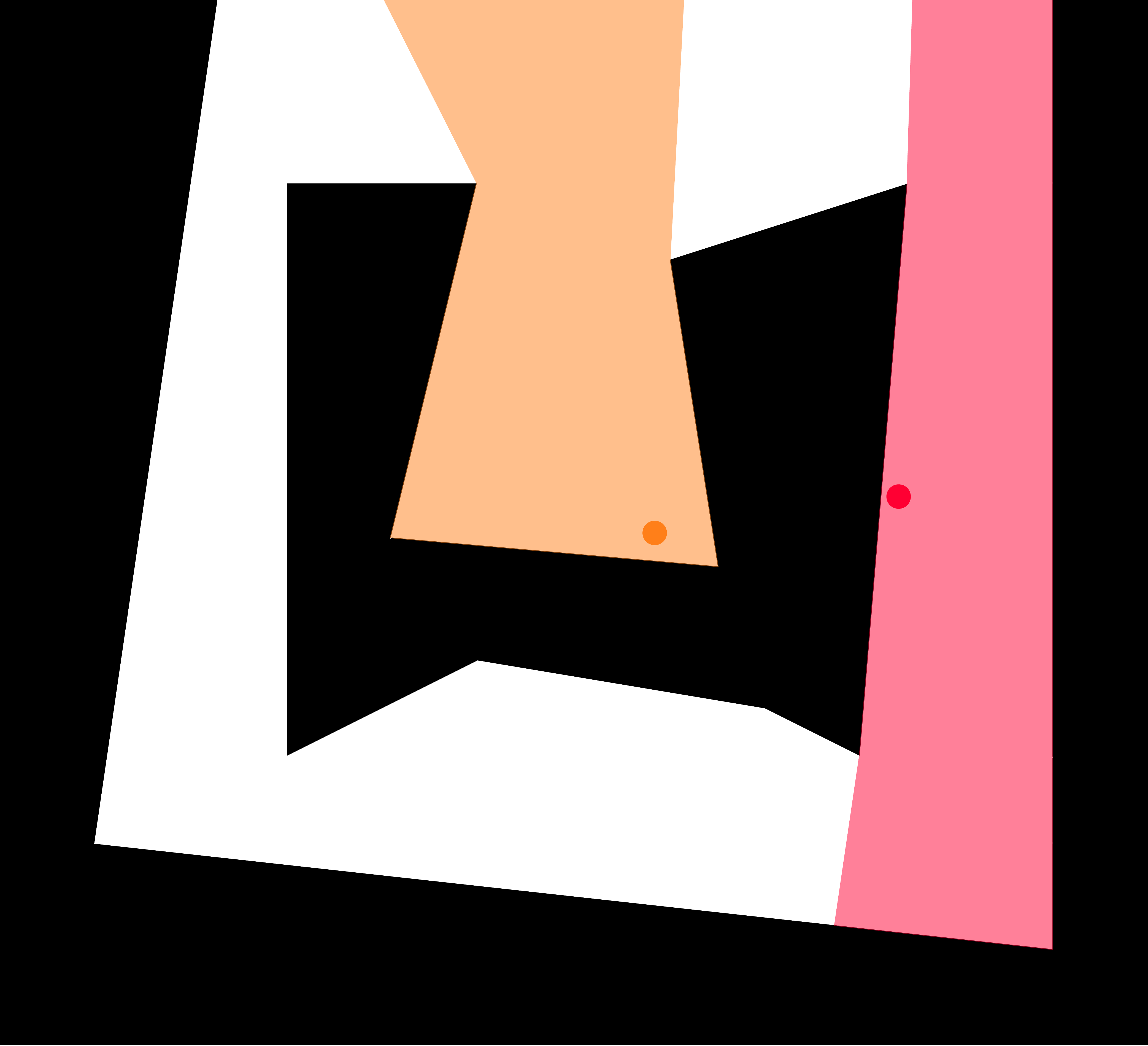}
    \caption{The first sample to be added. The \textcolor{blue}{$n^{\text{th}}$ pursuer} is removed (Algorithm~\ref{algo:junction}, line~\ref{line:jpc1}).}
  \end{subfigure}
  \medskip
  \begin{subfigure}[t]{.49\columnwidth}
    \centering
    \includegraphics[width=1\columnwidth]{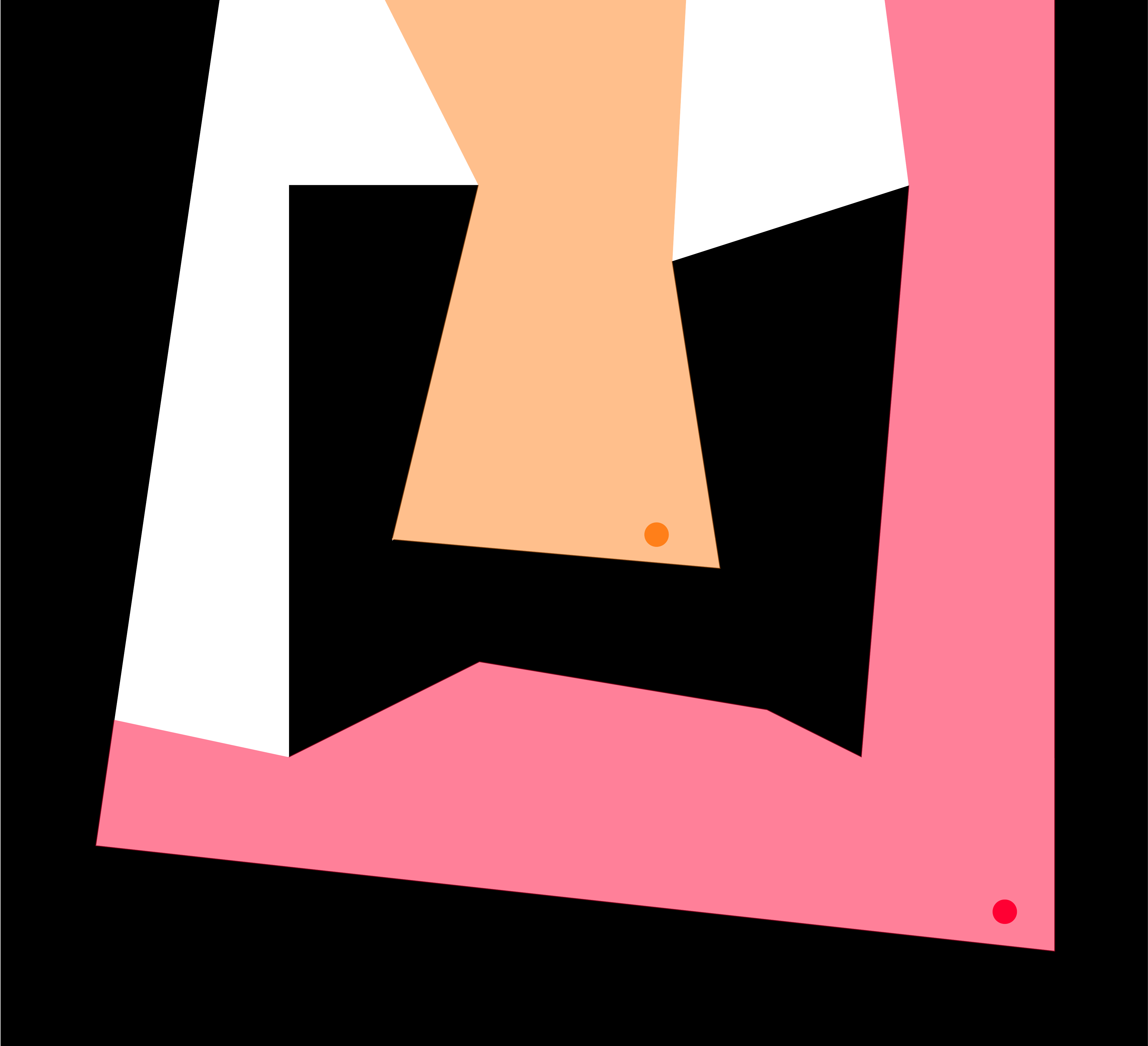}
    \caption{The second sample. The \textcolor{blue}{$n^{\text{th}}$ pursuer} is removed and the \textcolor{purple}{$i^{\text{th}}$ pursuer} moves to a random point in $V(\textcolor{purple}{p_i}) \cap V(\textcolor{blue}{p_n)}$. (Algorithm~\ref{algo:junction}, line~\ref{line:jpc3}).}
  \end{subfigure}
  \begin{subfigure}[t]{.49\columnwidth}
    \centering
    \includegraphics[width=1\columnwidth]{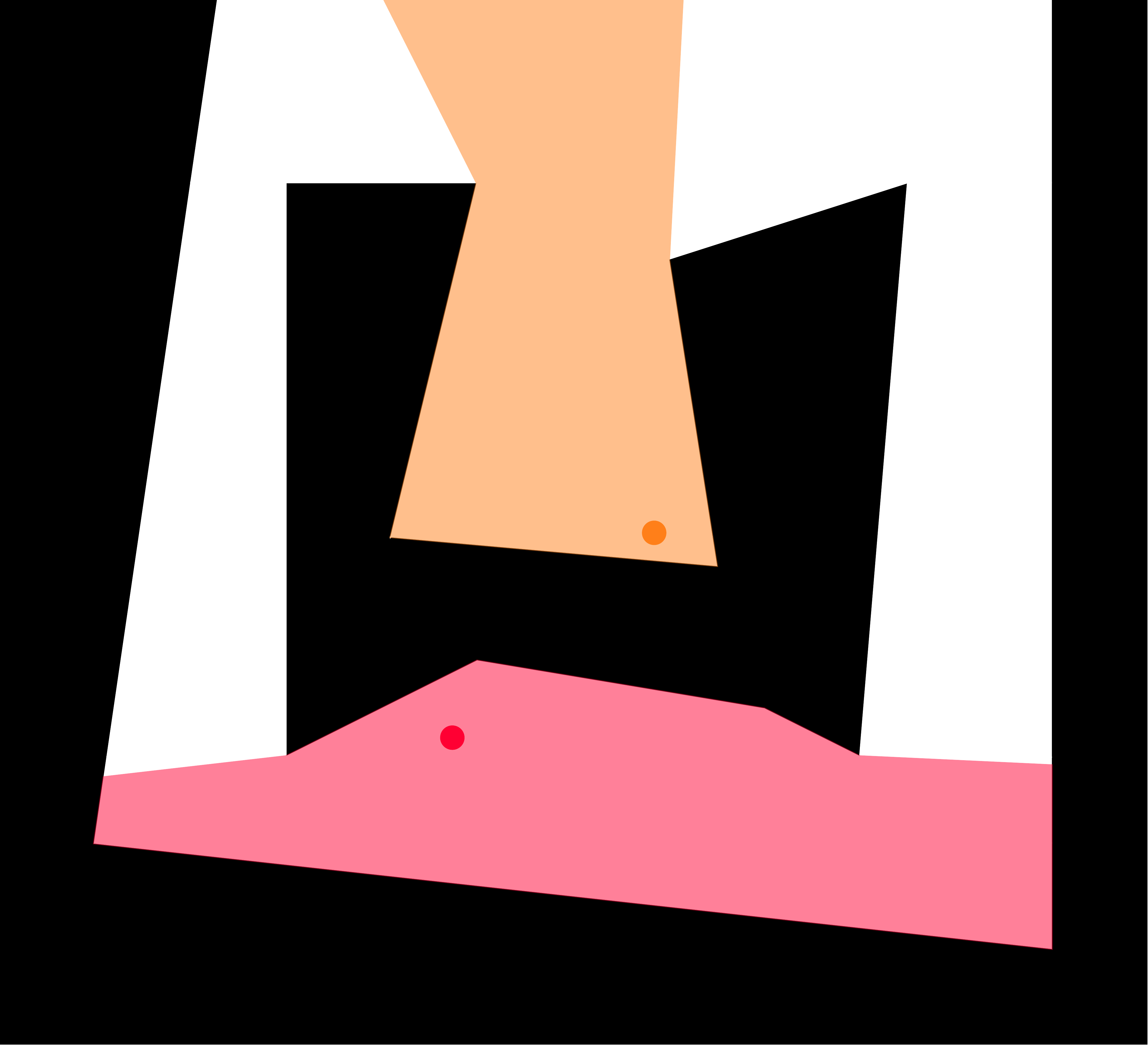}
    \caption{The third sample. The \textcolor{blue}{$n^{\text{th}}$ pursuer} is removed and the \textcolor{purple}{$i^{\text{th}}$ pursuer} takes its place. (Algorithm~\ref{algo:junction}, line~\ref{line:jpc2}).}
  \end{subfigure}
  \vspace{-5mm}
  \caption{An example of junction sampling.}
  \label{fig:junction}
\end{figure}

The objective in junction sampling is, informally, to add vertices and edges to the SG-PEG that allow remaining pursuers to `fill in' for the removed robot, wherever possible.  Figure~\ref{fig:junction} shows a simple example of a pursuer removed from a \jpc during \textsc{DropRobot}.  In this example, the lower pursuer is removed, leaving the bottom portion of the environment unobserved.  Junction sampling adds new samples that provide a path within the SG-PEG for the rightmost robot to visit the site of this lower portion.

This process, called \textsc{AddJunctionSamples}, is formalized in Algorithm~\ref{algo:junction}.  In the general case, the algorithm identifies a remaining pursuer at a position $p_i$ for which the visibility polygon intersects the visibility polygon of the position $p_n$ of the removed pursuer.  When this relationship is detected, we add a sample that places the $i^{\text{th}}$ pursuer in the intersection of the visibility polygons (see Figure~\ref{fig:junction}c) and another that places the $i^{\text{th}}$ pursuer at the former location of the $n^{\text{th}}$ pursuer (Figure~\ref{fig:junction}d).  This process is repeated for each $i$ and, via repeated calls to \textsc{AddJunctionSamples}, each step of the previous $k$-pursuer solution.

\subsubsection{Web sampling}
Though the structures introduced by junction sampling may be sufficient to build a SG-PEG that can generate a solution with $k-1$ pursuers, such success cannot be guaranteed.  Therefore, after exhausting the junction samples, Algorithm~\ref{algo:drop} continues with a broader sampling strategy called web sampling.  Web sampling was originally proposed for the failure-free version of the problem~\cite{OlsTum+21}.

\begin{figure}

  \begin{subfigure}[t]{.49\columnwidth}
    \centering
    \includegraphics[width=1\columnwidth]{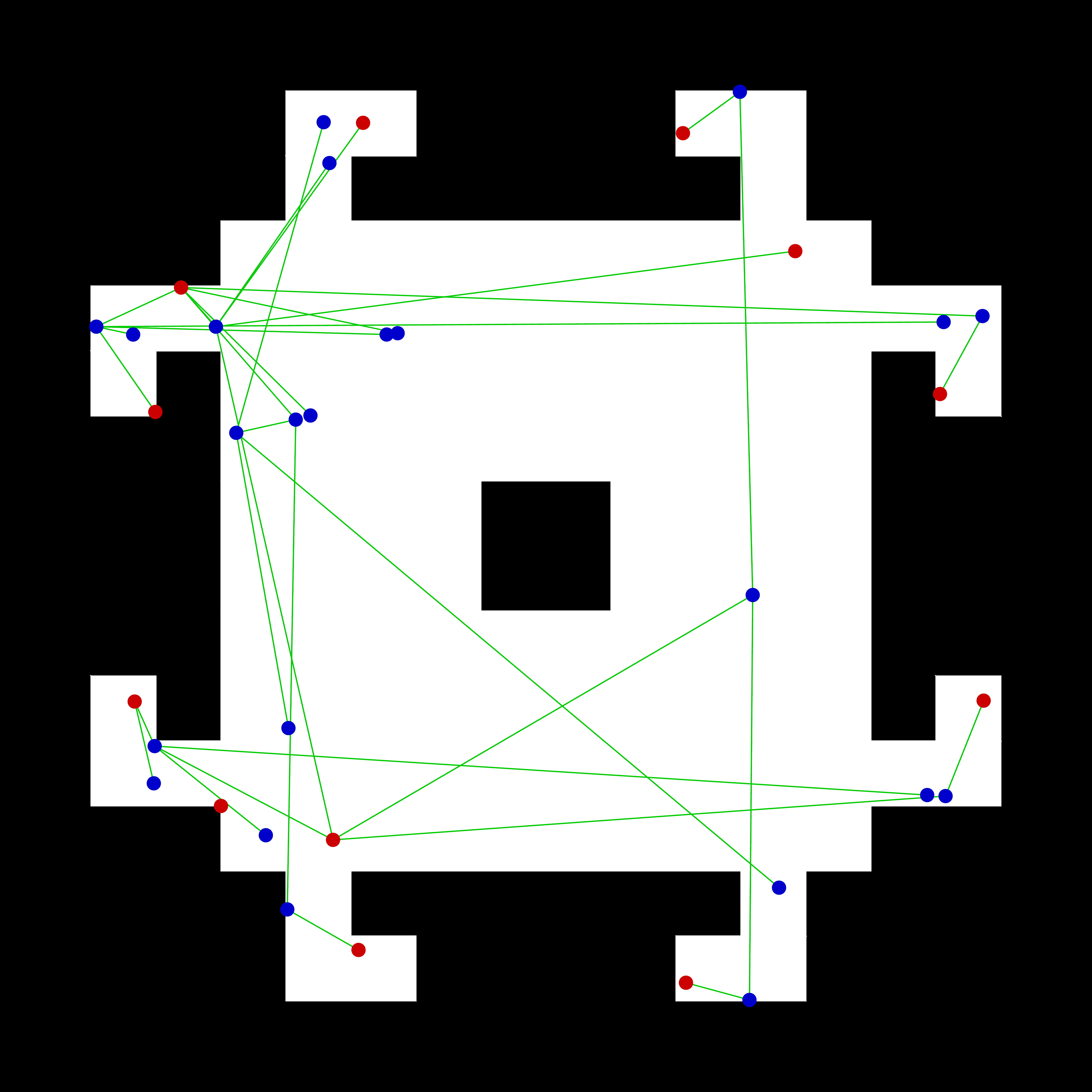}
    \caption{A set of 35 samples that form one complete web.}
  \end{subfigure}
  \begin{subfigure}[t]{.49\columnwidth}
    \centering
    \includegraphics[width=1\columnwidth]{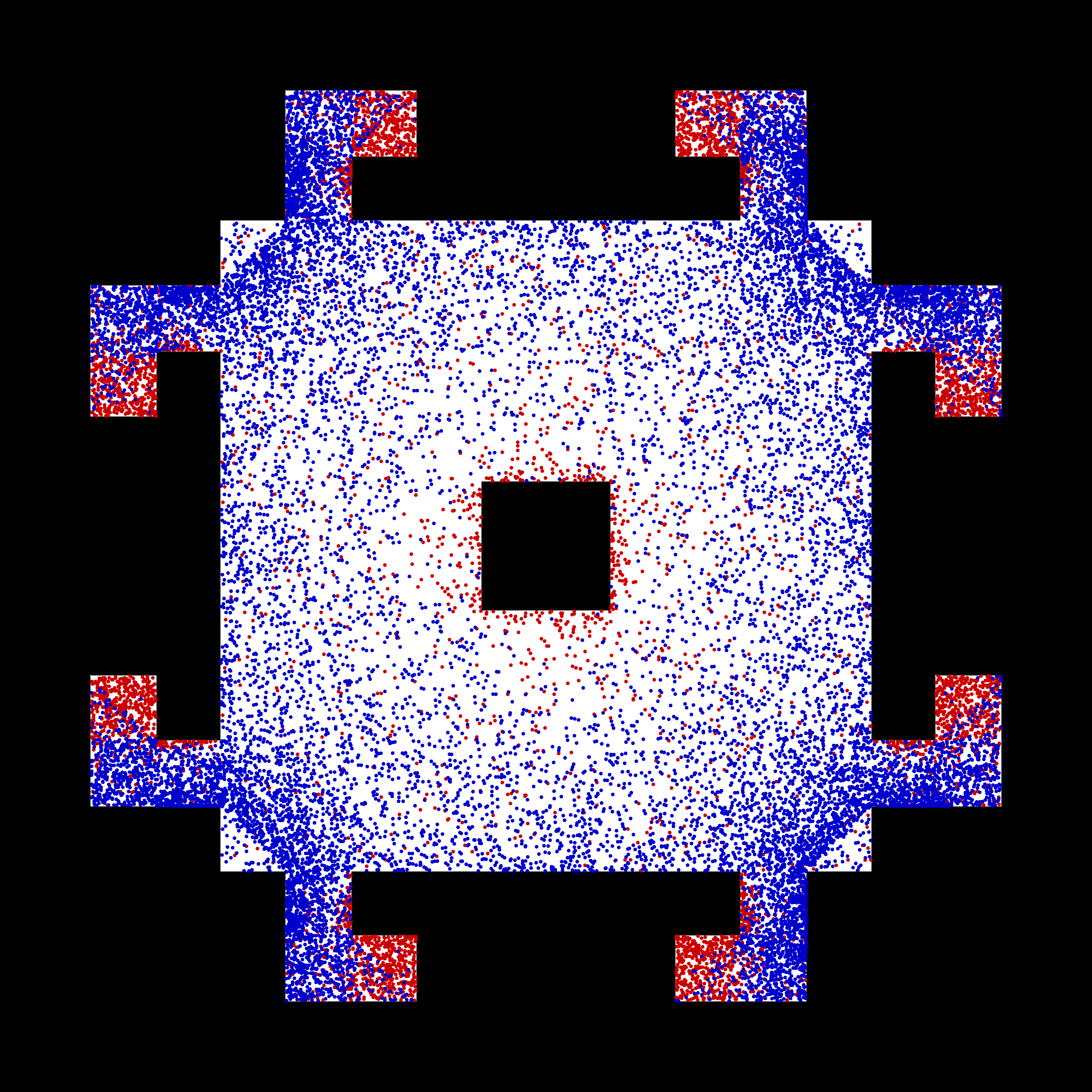}
    \caption{25000 samples drawn using web sampling. Notice how the points from $A$ (red) are biased towards the outer hooks, while the points from $B$ (blue) favor the regions connecting adjacent hooks.}
  \end{subfigure}
  \caption{An illustration of web sampling.}
  \label{fig:webs}
\end{figure}

The sampling approach is based on an underlying notion of a \emph{web}.  The intuition is select a collection of positions that can see the entire environment while also forming a connected graph via straight-line connections within $F$.
Generating a web occurs in two stages.
First, we draw a collection of points $A = \{ a_1, a_2, \dots , a_n \} \subset F$ which provide full visibility of the environment, i.e. $\bigcup_{1 \le i \le n} V(a_{i}) = F$.  This is done incrementally, by drawing samples from the unseen portion if $F$ until all of $F$ is seen by some point in $A$.
The second stage generates an intersection set $B$ as follows. For each pair of distinct points $a_i,a_j \in A$,  if $V(a_i) \cap V(a_j) \ne \emptyset$, we add a point $b \in V(a_i) \cap V(a_j)$ to $B$.  The combination $A \cup B$ forms one complete web; those points are utilized in a randomly shuffled order.  See Figure~\ref{fig:webs}.

To use these webs within \textsc{WebSample} (recall line~\ref{line:web} in Algorithm~\ref{algo:drop}), we generate one web for each of the $k-1$ pursuers.  Then select a random vertex $v$ from $G_{k-1}$ and, for two of the robots in that \jpc, form a new sample by replacing the existing positions with positions drawn (without replacement) from those pursuers' respective webs.  If any web ever has no more points to choose from, we generate new webs for each pursuer and continue the process. 

\subsection{Planning, execution, and replanning}
Armed with the \textsc{DropRobot} method, we can consider how to use that algorithm for the overall problem.

\subsubsection{Generating the initial solution}\label{sec:initial}
To begin, we must generate an initial solution that the full complement of $n$ robots can begin to execute.  First, we generate a trivial solution, namely a strategy where no movement is required by the pursuers because their visibility polygons fully cover the environment.  We do so by iteratively adding pursuers at random unseen locations until no shadows remain.  This single \jpc becomes our trivial solution.

Note, however ---recalling that only $n$ robots are available at the start---, that it is rather likely that the trivial solution will require more than $n$ robots.  If so, we repeatedly apply \textsc{DropRobot}, selecting the pursuer to remove at random, until a solution requiring only $n$ pursuers has been formed.  The pursuer team then begins to execute this strategy.\footnote{It is possible in principle that the trivial solution may require $n$ robots or less.  In that case, we can ignore any additional robots beyond the $m$ that are required for the trivial solution and simply `execute' the trivial solution.}

\subsubsection{Replanning after pursuer failures}\label{sec:replan}
If, during the execution of the search, a pursuer fails for some reason, a replanning operation is required.  In that case, we pause the pursuers' movement until a new solution with one fewer pursuer is generated.  This new solution may be generated directly by \textsc{DropRobot}.  Notice that the inputs to that algorithm include the current state of the search (including the current \jpc and the current shadow label), which are leveraged to replan more rapidly than planning from scratch each time.  Once a new solution is computed, the pursuers resume their search.

\newcommand{\spider}{Figure~\ref{fig:webs}\xspace}
\newcommand{\henv}{Figure~\ref{fig:shadows}\xspace}
\newcommand{\nineroom}{Figure~\ref{fig:notation}\xspace}

\section{Evaluation}\label{sec:experiments}
\begin{figure*}
  \begin{subfigure}[t]{.19\textwidth}
    \centering
    \includegraphics[width=1.2\columnwidth, angle=90]{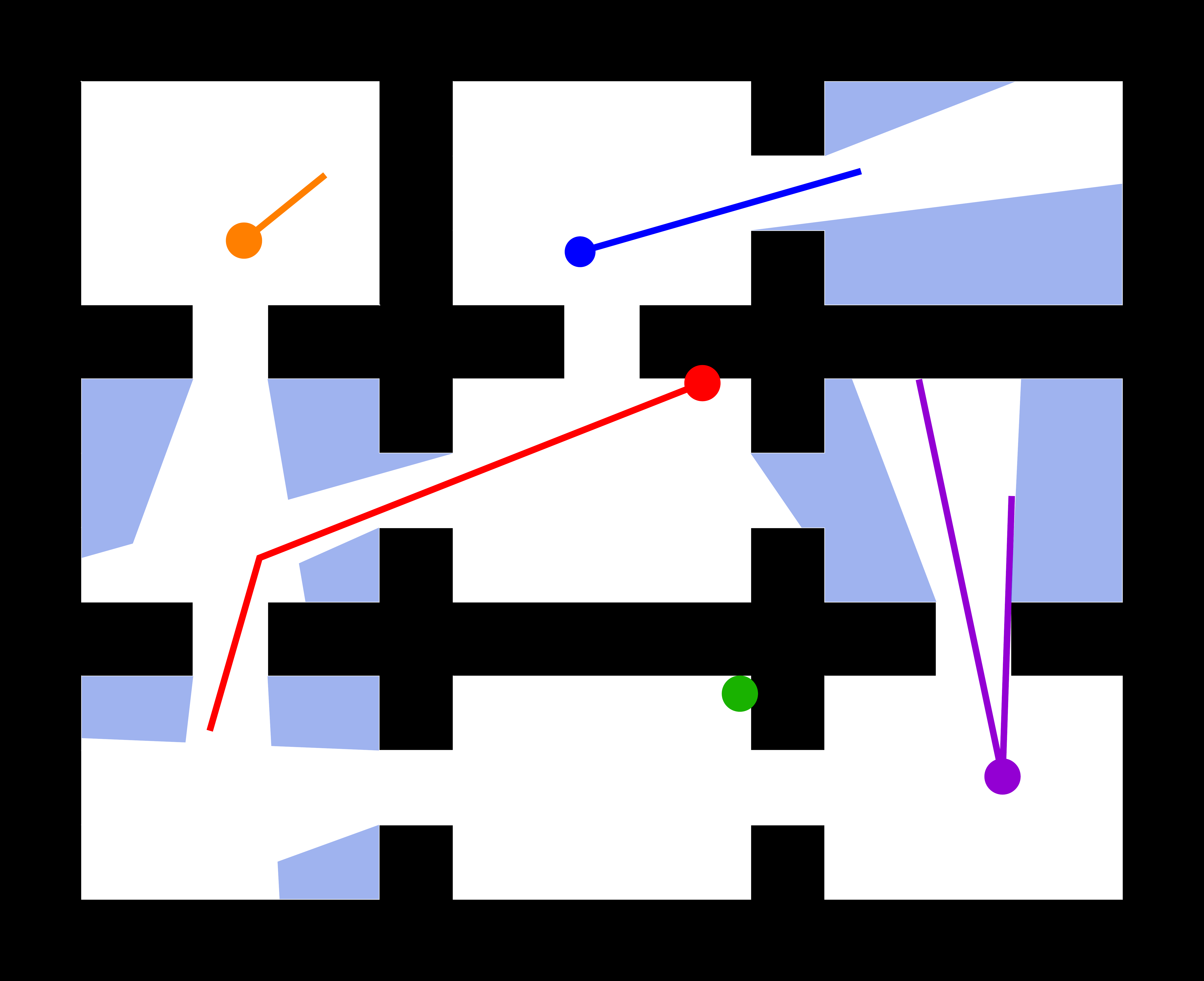}
    \caption{An initial solution with 5 pursuers.}
  \end{subfigure}
  \begin{subfigure}[t]{.19\textwidth}
    \centering
    \includegraphics[width=1.2\columnwidth, angle=90]{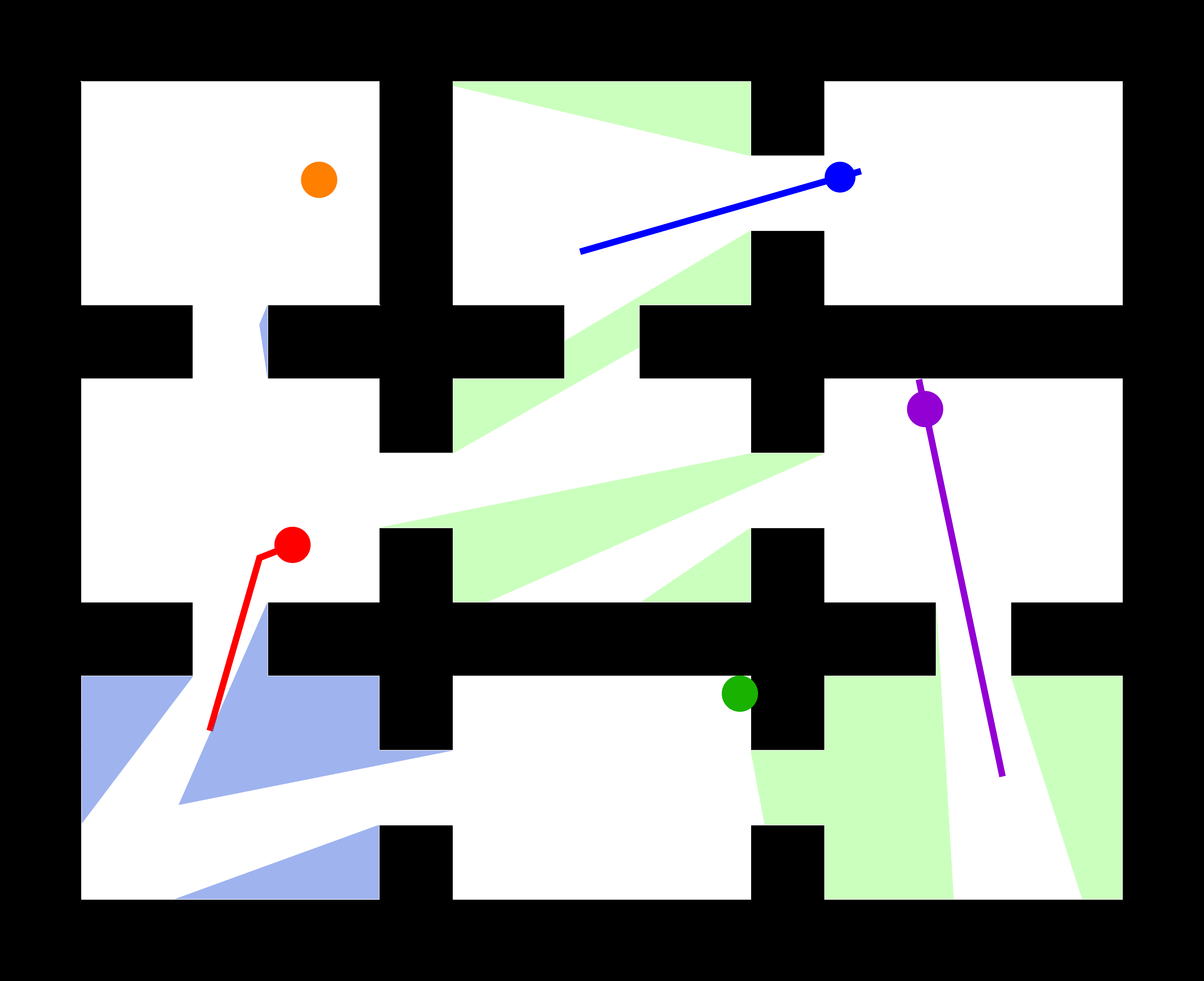}
    \caption{The problem state right before the green pursuer fails.}
  \end{subfigure}
\begin{subfigure}[t]{.19\textwidth}
    \centering
    \includegraphics[width=1.2\columnwidth, angle=90]{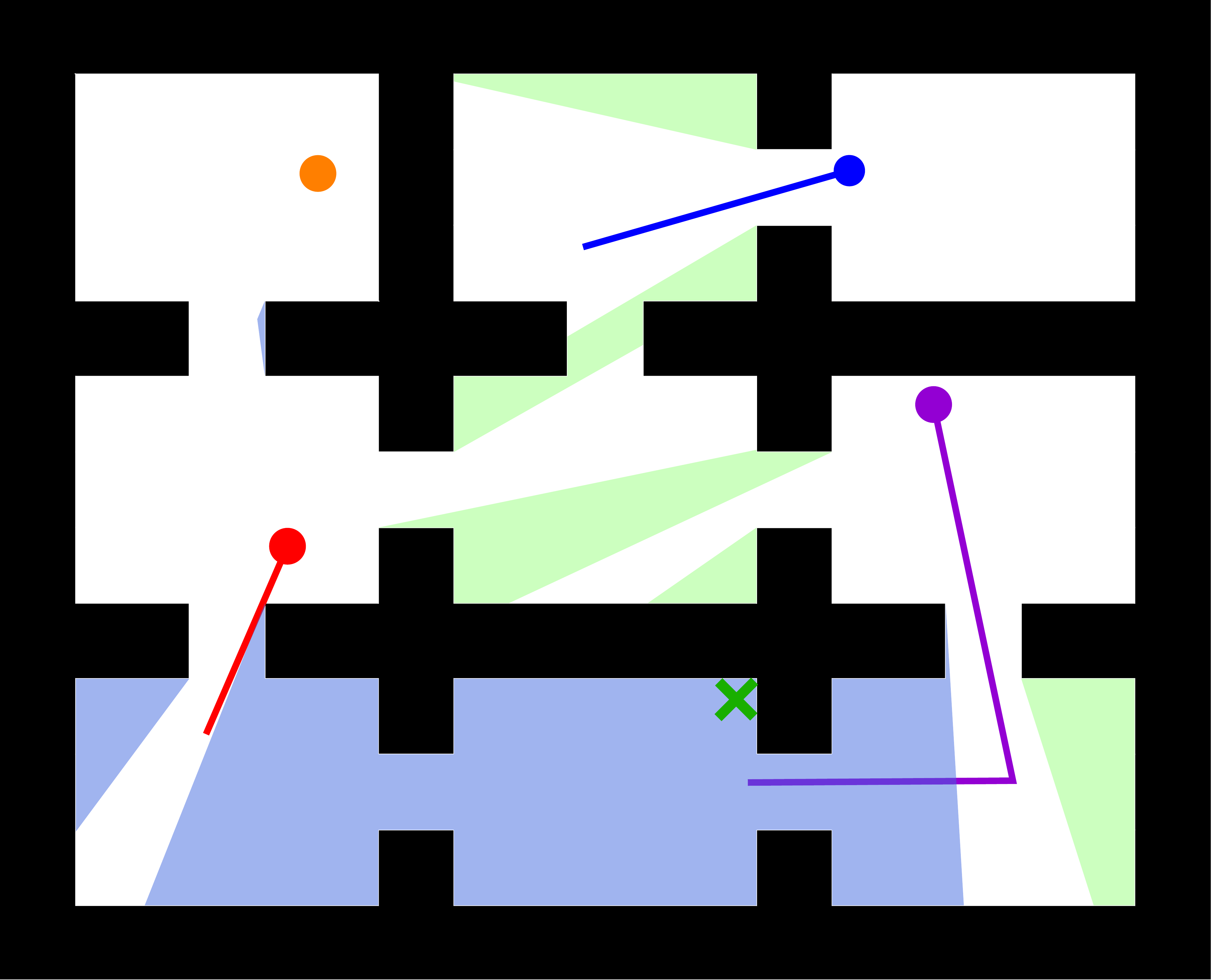}
    \caption{The new solution paths generated after the green pursuer fails.}
  \end{subfigure}
  \begin{subfigure}[t]{.19\textwidth}
    \centering
    \includegraphics[width=1.2\columnwidth, angle=90]{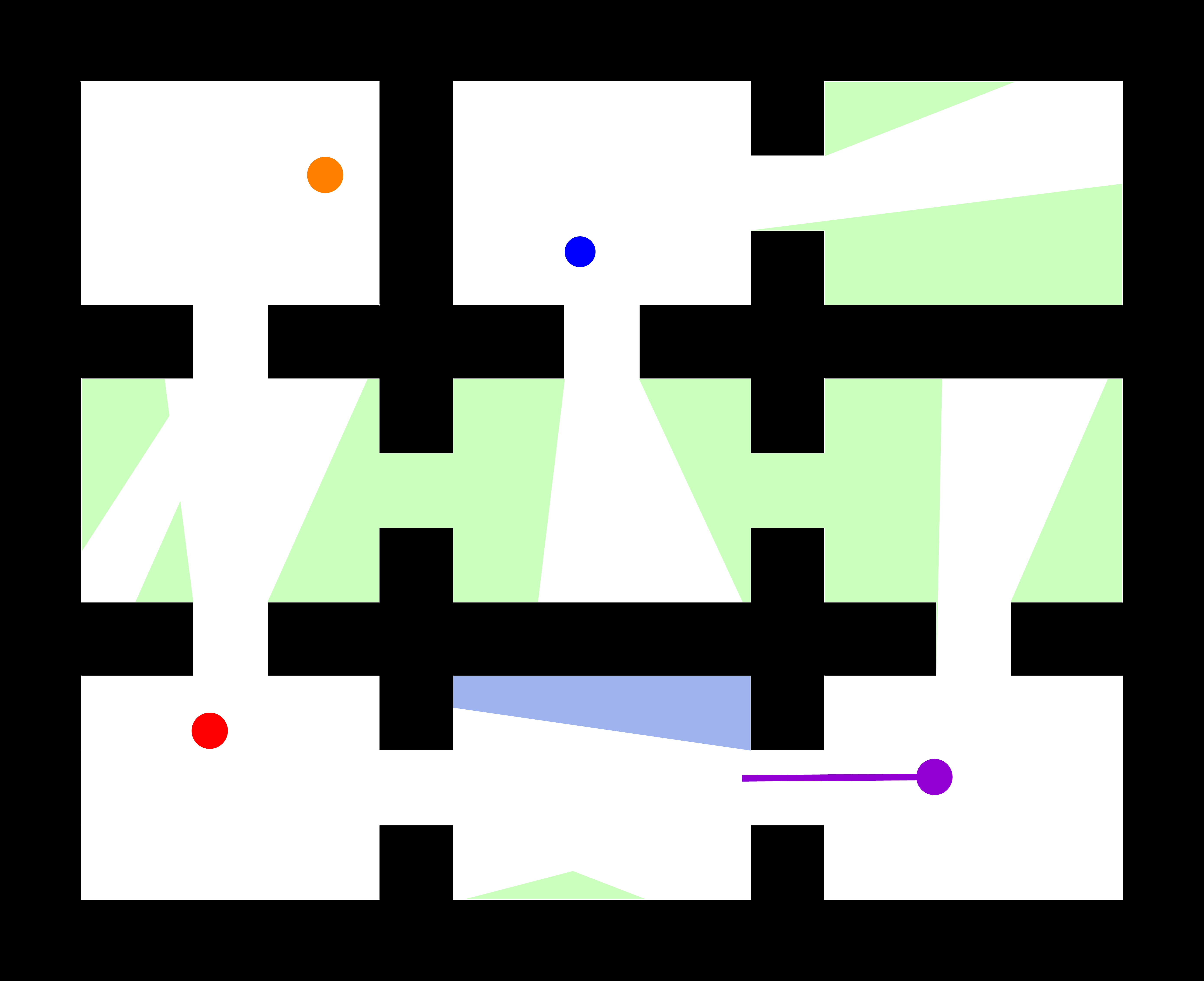}
    \caption{The problem state right before the orange pursuer fails.}
  \end{subfigure}
  \begin{subfigure}[t]{.19\textwidth}
    \centering
    \includegraphics[width=1.2\columnwidth, angle=90]{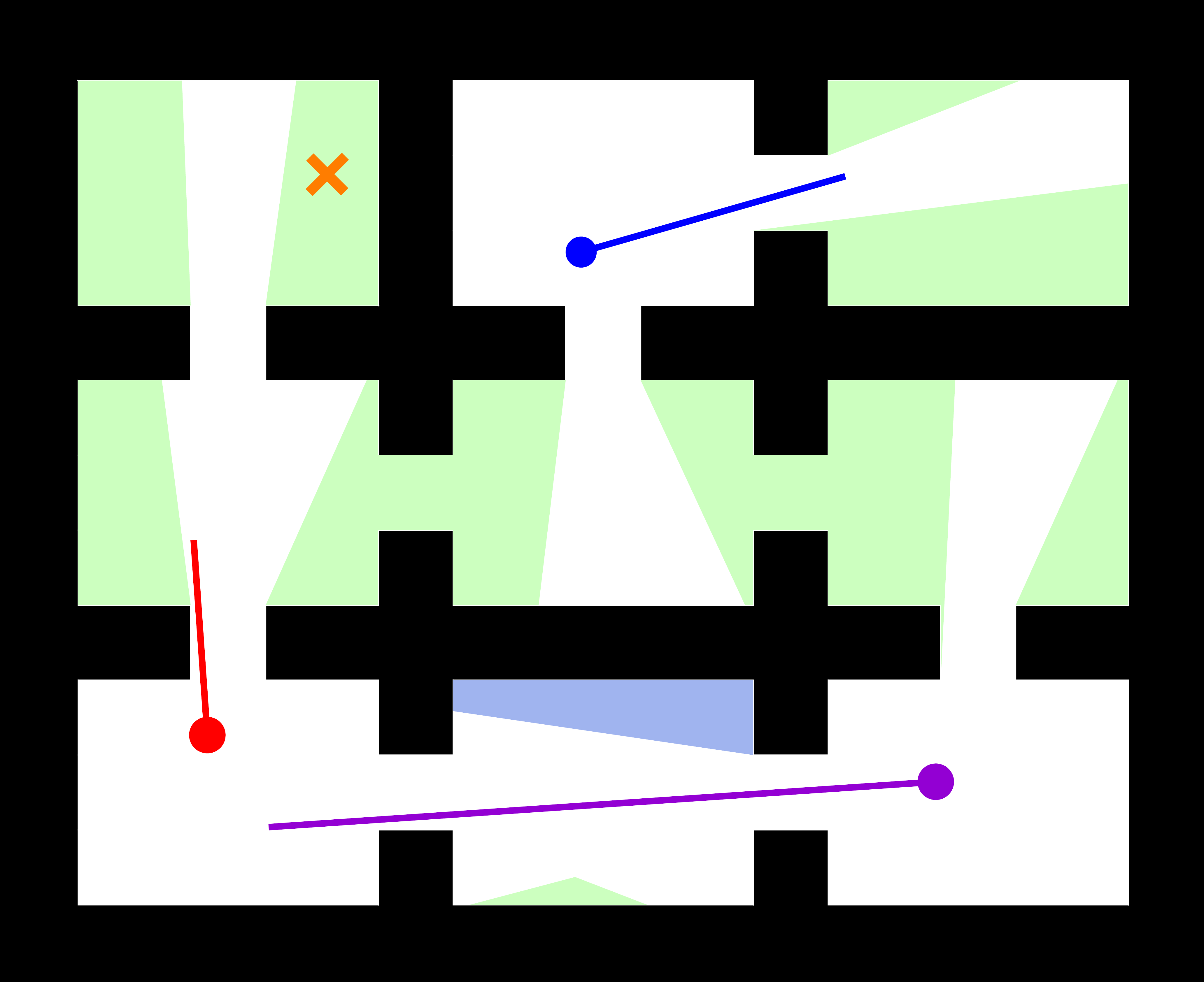}
    \caption{The new solution paths generated after the orange pursuer fails.}
  \end{subfigure}
  \caption{Snapshots of our algorithm through a single successful execution ($n = 5, m = 2$).}
  \label{fig:sols}
\end{figure*}

We implemented our algorithm in C++ and executed the code on an Ubuntu 20.04 laptop equipped with an Intel i7-10510U CPU and 16GB of RAM. %

An example execution is illustrated in Figure~\ref{fig:sols}. First, an initial solution is generated (Figure~\ref{fig:sols}a). Next, Figures~\ref{fig:sols}b,c represent the input and output of Algorithm~\ref{algo:drop} when the green pursuer malfunctions. Similarly, Figures~\ref{fig:sols}d,e show the state before and after the failure of the orange pursuer.

We simulated teams initially consisting of $n=5$ pursuers\footnote{Increasing $n$ has a positive effect on the planning time of the proposed algorithm, since, by construction, we need to generate solutions for each number of pursuers between the number of pursuer in the trivial solution and $n$. In contrast, OTSO21 can struggle with larger values of $n$ due to the increased complexity of the joint configuration space, making it more difficult to connect pursuer configurations. Thus, we hold $n$ fixed at 5 to enable a fair comparison.} in three different environments, depicted in  Figures~\ref{fig:notation}, \ref{fig:shadows}, and \ref{fig:webs}. These environments were selected because they highlight several interesting attributes, such as hard to reach corners, narrow corridors, and evenly spaced obstacles. Additionally, these environments allow us to more directly compare against existing results. In particular, we compare the algorithm presented in Section~\ref{sec:ad} (`this paper') against our previous algorithm~\cite{OlsTum+21} (`OTSO21'), which was designed for the failure-free setting, as a baseline.
During each execution, we simulated $m$ pursuer failures.  For each failure, a randomly-selected pursuer was removed when the pursuers had completed a percentage $\beta$ of their planned paths.  For OTSO21, the algorithm was executed from scratch for the initial solution and at each robot failure.  Runs were conducted for all four combinations of $m \in \{ 2, 3 \}$ and $\beta \in \{ 30\%, 70\%\}$.

Each trial was limited to at most 10 minutes of run time, including both planning time and (simulated) execution time.
If, after that time, the robots had not yet successfully cleared all shadows, the simulation would have been considered a failure.  In the results presented here, none of the trials failed.

For each combination of environment, algorithm, team size $n$, number of failures $m$, and failure time $\beta$, we conducted 25 trials.  The success or failure of the run and total computation time spent planning and replanning were recorded.  Planning time is summarized by the mean $(\mu)$ and the standard deviation $(\sigma)$ over all trials.
Tables~\ref{tab:simulations} and \ref{tab:simulations2} report the results, from which
a few conclusions may be drawn.

\newcommand{\timpaper}{OTSO21\xspace}
\newcommand{\thing}[1]{\par\bigskip\noindent\emph{#1}\quad}

    \thing{Replanning is beneficial} Recall from Section~\ref{ssect:js} that \sample{} was developed to
        ``recover'' information in the event of a pursuer failure. 
        The notable improvements for the proposed algorithm compared to \timpaper in the environments of \nineroom and \henv can be attributed to efficiencies gained by re-planning rather than starting from scratch.
        In the environment of Figure~\ref{fig:webs}, the proposed algorithm performed similarly to OTSO21, likely due to the complexity of the environment resulting in a high number of pursuers in its trivial solutions and subsequently more calls to \textsc{DropRobot} to reach the initial solution.

    \thing{Later failures are easier to recover} For the trials with $\beta = 70\%$, the total planning time was less than when $\beta=30\%$. This is likely due to the fact that allowing more time to traverse the solution path will, in many cases, provide the next planning stage with an improved shadow label (i.e. more cleared shadows), reducing the difficulty of the replanning problem.
    
    \thing{Impacts of the number of failures}
    Increasing from $m=2$ to $m=3$ increased the planning time for both algorithms.  We speculate that this can be attributed to the additional pursuer failure
    for which both the proposed algorithm and \timpaper are required to recompute strategies.

\begin{table}[t]
    \caption{Simulation results for $n=5$ initial pursuers and $m=2$ failures.}
    \label{tab:simulations}
    \vspace{-2mm}
    \centering
    \resizebox{0.98\columnwidth}{!}{
    \begin{tabular}{lcr@{\hspace{0.75\tabcolsep}}rr@{\hspace{0.75\tabcolsep}}rr@{\hspace{0.75\tabcolsep}}r}
                    & success & \multicolumn{2}{c}{planning time (s)} \\
                    & \multicolumn{1}{c}{rate} &  \multicolumn{1}{c}{$\mu$} & \multicolumn{1}{c}{$\sigma$} \\ \hline
        \multicolumn{1}{c}{\textbf{\nineroom}}\\ \hline
        This paper ($\beta = 30\%$)             & 100\% & 46.09 & 21.50  \\
        OTSO21 \hspace{1.25mm} ($\beta = 30\%$) & 100\% & 99.57 & 65.03  \\
        This paper ($\beta = 70\%$)             & 100\% & 32.46 & 14.84  \\ 
        OTSO21 \hspace{1.25mm} ($\beta = 70\%$) & 100\% & 89.36 & 59.62  \\ \hline
         \multicolumn{1}{c}{\textbf{\henv}}\\ \hline
        This paper ($\beta = 30\%$)             & 100\% & 6.47  & 7.46  \\
        OTSO21 \hspace{1.25mm} ($\beta = 30\%$) & 100\% & 56.13 & 17.20 \\
        This paper ($\beta = 70\%$)             & 100\% & 5.07  & 5.97  \\ 
        OTSO21 \hspace{1.25mm} ($\beta = 70\%$) & 100\% & 52.87 & 17.59 \\ \hline
        \multicolumn{1}{c}{\textbf{\spider}}\\ \hline
        This paper ($\beta = 30\%$)             & 100\% & 89.13 & 39.08  \\
        OTSO21 \hspace{1.25mm} ($\beta = 30\%$) & 100\% & 94.79 & 27.95  \\
        This paper ($\beta = 70\%$)             & 100\% & 72.06 & 34.17  \\
        OTSO21 \hspace{1.25mm} ($\beta = 70\%$) & 100\% & 73.98 & 20.56  \\ \hline
    \end{tabular}
    }
\end{table}

\begin{table}[t]
    \caption{Simulation results for $n=5$ initial pursuers and $m=3$ failures.}
    \vspace{-2mm}
    \centering
    \resizebox{0.98\columnwidth}{!}{
    \begin{tabular}{lcr@{\hspace{0.75\tabcolsep}}rr@{\hspace{0.75\tabcolsep}}rr@{\hspace{0.75\tabcolsep}}r}
                  & success                  & \multicolumn{2}{c}{planning time (s)} \\
                  & \multicolumn{1}{c}{rate} &  \multicolumn{1}{c}{$\mu$} & \multicolumn{1}{c}{$\sigma$} \\ \hline
        \multicolumn{1}{c}{\textbf{\nineroom}}\\ \hline
        This paper ($\beta = 30\%$)             & 100\% & 63.92  & 29.89  \\
        OTSO21 \hspace{1.25mm} ($\beta = 30\%$) & 100\% & 117.90 & 64.78  \\
        This paper ($\beta = 70\%$)             & 100\% & 39.88  & 20.61  \\ 
        OTSO21 \hspace{1.25mm} ($\beta = 70\%$) & 100\% & 102.57 & 63.03  \\ \hline
        \multicolumn{1}{c}{\textbf{\henv}}\\ \hline
        This paper ($\beta = 30\%$)             & 100\% & 8.77  & 8.20  \\
        OTSO21 \hspace{1.25mm} ($\beta = 30\%$) & 100\% & 58.35 & 16.60 \\
        This paper ($\beta = 70\%$)             & 100\% & 6.54  & 6.41  \\ 
        OTSO21 \hspace{1.25mm} ($\beta = 70\%$) & 100\% & 53.78 & 16.64 \\ \hline
        \multicolumn{1}{c}{\textbf{\spider}}\\ \hline
        This paper ($\beta = 30\%$)             & 100\% & 112.19 & 39.02  \\
        OTSO21 \hspace{1.25mm} ($\beta = 30\%$) & 100\% & 108.11 & 30.13  \\
        This paper ($\beta = 70\%$)             & 100\% & 74.81  & 35.71  \\
        OTSO21 \hspace{1.25mm} ($\beta = 70\%$) & 100\% & 77.03  & 19.77  \\ \hline
    \end{tabular}
    }
    \label{tab:simulations2}
\end{table}

\section{Conclusion}\label{sec:conclusion} 
We presented a method of deconstructing higher dimensional solutions in order to alleviate the issue of potential robotic failures in a visibility-based pursuit-evasion problem. We did this by building a new sampling strategy which allowed us to utilize previously computed information. Our algorithm was able to greatly our-perform existing algorithms in the context of our problem. 
Future work could include generating solutions that are intentionally robust to failures. That is, a solution that would still be a solution if a limited number of pursuers were to completely malfunction. One possible approach to this problem would be to expand the shadow labels from single clear/contaminated bits, to a richer representation of the sets of pursuer failures under which that shadow would nonetheless be clear.

\newrefcontext[sorting=nyt]
\printbibliography

\end{document}